\documentclass{article}
\usepackage{authblk}
\usepackage[utf8]{inputenc} 
\usepackage[T1]{fontenc}
\usepackage[a4paper, margin=2.7cm]{geometry}
\usepackage{hyperref}
\usepackage{tgbonum}

\usepackage{multirow}
\usepackage{pifont}
\usepackage{adjustbox}
\usepackage{caption}
\usepackage{subcaption}
\usepackage{amsmath}
\usepackage{wrapfig}

\usepackage{url}            
\usepackage{booktabs}       
\usepackage{amsfonts}       
\usepackage{nicefrac}       
\usepackage{microtype}      
\usepackage{xcolor}         

\begin{document}
\title{Universal Domain Adaptation from Foundation Models: A Baseline Study}

\author[ ]{Bin Deng}
\author[ ]{Kui Jia}
\affil[ ]{South China University of Technology}
\affil[ ]{\normalsize\textit{eebindeng@mail.scut.edu.cn}}

\maketitle
\begin{abstract}
Foundation models (e.g., CLIP or DINOv2) have shown their impressive learning and transfer capabilities in a wide range of visual tasks, by training on a large corpus of data and adapting to specific downstream tasks. It is, however, interesting that foundation models have not been fully explored for universal domain adaptation (UniDA), which is to learn models using labeled data in a source domain and unlabeled data in a target one, such that the learned models can successfully adapt to the target data.
In this paper, we make comprehensive empirical studies of state-of-the-art UniDA methods using foundation models.
We first observe that, unlike fine-tuning from ImageNet pre-trained models, as previous methods do, fine-tuning from foundation models yields significantly poorer results, sometimes even worse than training from scratch.
While freezing the backbones, we demonstrate that although the foundation models greatly improve the performance of the baseline method that trains the models on the source data alone, existing UniDA methods generally fail to improve over the baseline. This suggests that new research efforts are very necessary for UniDA using foundation models.
Based on these findings, we introduce \textit{CLIP distillation}, a parameter-free method specifically designed to distill target knowledge from CLIP models. The core of our \textit{CLIP distillation} lies in a self-calibration technique for automatic temperature scaling, a feature that significantly enhances the baseline's out-class detection capability. Although simple, our method outperforms previous approaches in most benchmark tasks, excelling in evaluation metrics including H-score/H$^3$-score and the newly proposed universal classification rate (UCR) metric. We hope that our investigation and the proposed simple framework can serve as a strong baseline to facilitate future studies in this field.
The code is available at \url{https://github.com/szubing/uniood}.
\end{abstract}

\section{Introduction}
A foundational goal of machine visual is to develop a model that can be applied to data from different distributions. With the emergence of many large-scale pre-trained models such as CLIP \cite{radford2021learning}, ALIGN \cite{jia2021scaling}, and DINOv2 \cite{oquab2023dinov2}, significant progress has been made recently towards achieving this goal. These "foundation models" \cite{bommasani2021opportunities} often exhibit significantly greater robustness to various benchmark distribution shifts compared to standard training models. Taking image classification as an example, both CLIP and a standard ResNet50 achieve an accuracy of 76\% on ImageNet, but CLIP has shown significant improvements with an accuracy increase of 6\% on ImageNetV2 and an accuracy increase of 35\% on ImageNet Sketch \cite{radford2021learning}. Due to the powerful ability of these foundation models, techniques for applying them on downstream applications are increasingly important. Indeed, the research community has spent significant effort during the past few years to improving the fine-tuning of these models for various downstream tasks, including few-shot classification \cite{lin2023multimodality}, out-of-distribution (OOD) detection \cite{esmaeilpour2022zero}, and OOD generalization \cite{wortsman2022robust, lee2023surgical}, among others.

Surprisingly, universal domain adaptation (UniDA) \cite{you2019universal}, one of the practical applications that aims to adapt to one specific target domain without any restriction on label sets, has not been thoroughly explored to date under the powerful foundation models. This paper aims to fill this gap by initially assessing the performance of state-of-the-art UniDA methods when applied to foundation models. Through comprehensive experiments, we conclude several interesting findings. First of all, unlike fine-tuning from pre-trained models on ImageNet, fine-tuning from foundation models yields significantly poorer results, sometimes even worse than training from scratch. We then freeze the foundation models and focus solely on updating the classifier head. In this scenario, all methods achieved substantial improvements over their prior results that are reliant on ImageNet pre-trained models. However, the performance gap between the Source Only (SO) baseline and state-of-the-art (SOTA) methods has notably narrowed, rendering them largely comparable across various benchmark tasks. These findings suggest that new research efforts are very necessary for UniDA using foundation models.

Based on our empirical observations, we present \textit{CLIP distillation}, a parameter-free technique designed to distill knowledge from CLIP models. The core of our \textit{CLIP distillation} method revolves around a self-calibration approach that automatically adjusts temperature scaling. This feature significantly enhances the baseline model's ability to detect out-class samples. Additionally, we introduce a new evaluation metric for UniDA that is threshold- and ratio-free, making it suitable for methods that do not consider threshold effects.
Despite its simplicity, our method demonstrates exceptional robustness and efficacy in various task scenarios, spanning open-partial, open, closed, and partial UniDA settings. It excels in both the established H-score/H$^3$-score metrics and the novel UCR metric. This straightforward approach sets a new standard for UniDA using foundation models, providing a solid baseline for future research in this field.

The main contributions of this paper are summarized as follows:
\begin{itemize}
    \item To the best of our knowledge, we are the first to tackle the UniDA challenge and conduct comprehensive studies into existing methods when applied to foundation models. Our findings underscore the urgent need for further research for UniDA using these powerful foundation models.
    \item We propose \textit{CLIP distillation} for UniDA, which sets a new baseline for adaptation from foundation models. Our method includes a self-calibration technique for automatic temperature scaling, rendering \textit{CLIP distillation} parameter-free and robust across diverse task settings.
    \item  We propose a novel evaluation metric for UniDA, the Universal Classification Rate (UCR), which is insensitive to threshold and ratio considerations. Additionally, in order to facilitate rigorous and replicable experimentation in UniDA, we have developed and made available the UniOOD framework. UniOOD simplifies the incorporation of new datasets and algorithms through a few lines of code, ensuring fairer comparisons between various methods.
\end{itemize}

\section{Related works}
\noindent \textbf{Universal domain adaptation.} Different from the traditional domain adaptation (DA) problem, which assumes all labels in the target domain are identical to the source domain, universal domain adaptation (UniDA) \cite{you2019universal} assumes that there is no prior knowledge about the label relationship between source and target domains. Due to the existence of labels shift in UniDA, classical DA methods of adversarial adaptation such as DANN \cite{ganin2016domain} often suffer from negative transfer. To address this problem, UAN \cite{you2019universal} and CMU \cite{fu2020learning} use sample-level uncertainty criteria to assign weights for each sample before adversarial alignment. In addition to adversarial adaptation, self-training or self-supervised-based methods usually have better performance due to the exploiting of discriminative representation on the target domain. Among these, DANCE \cite{saito2020universal} uses self-supervised neighborhood clustering to learn the target data structure; DCC \cite{li2021domain} exploits cross-domain consensus knowledge to discover discriminative clusters of both domains; MATHS \cite{chen2022mutual} designs a contrastive learning scheme to nearest neighbors for feature alignment; OVANet \cite{saito2021ovanet} proposes to train a one-vs-all classifier for each class and applies entropy minimization to target samples during adaptation; and more recently, UniOT \cite{chang2022unified} uses optimal transport criteria to select more confident clusters to target samples for self-training. However, all of these methods are evaluated solely using models pre-trained in ImageNet. In this paper, we compare against the most state-of-the-art methods under the foundation models. To our knowledge, these methods are DANCE, OVANet, and UniOT, which we detail in Section \ref{section:methods-review} respectively. We show that there exists a strong baseline that can be competitive with or outperform the more complex methods listed above when using foundation models.

\noindent \textbf{Adaptation of foundation models.} The exceptional performance of foundation models in traditional vision tasks has led to a growing interest in developing more effective adaptive methods. In addition to adopting linear probing \cite{he2022masked}, full fine-tuning \cite{neyshabur2020being}, or zero shot \cite{radford2021learning} to the backbone models, many new strategies or methods have been proposed. For example, prompt learning based methods \cite{zhou2022learning, zhou2022conditional, gu2023towards} propose to learn better prompts under the language-vision models. CLIP-Adapter \cite{gao2021clip} and Tip-Adapter \cite{zhang2022tip} are going to construct additional light models for efficient fine-tuning while freezing the backbone models. Surgical fine-tuning \cite{lee2023surgical} suggests selectively fine-tuning a subset of layers based on different types of distribution shift. WiSE-FT \cite{wortsman2022robust} proposes to enhance the model robustness by integrating the zero-shot model and the fine-tuning model. And more recently, cross-model adaptation \cite{lin2023multimodality} shows the most powerful few-shot ability to CLIP based models by incorporating multi-modalities as training samples for ensemble training. In this paper, different from all these methods that aim to adapt models for closed-set classification task, we exploit to adapt for UniDA problem. We also show how effective would be if these representative methods are directly applied for the UniDA tasks.

\noindent \textbf{Related subfields.} UniDA is also closely related to open-set recognition (OSR) \cite{scheirer2012toward} and out-of-distribution (OOD) detection \cite{hendrycks2016baseline}. OSR extends the closed-set classification to a more realistic open-set classification, where test samples may come from domains of unknown classes. This setting is very similar to UniDA but it assumes that there exists no domain shift and that one can not access the target domain during training. OOD detection, on the other hand, focuses on detecting the out-class samples only. In theory, a recent work of \cite{fangout2022} unifies OSR and OOD detection into the same framework and shows that the loss criterion must be carefully designed otherwise it may face an intractable learning problem. In this paper, we are inspired by these works and introduce a similar evaluation metric of UCR for UniDA. Although many methods of OSR and OOD detection have been proposed during the past few years, a recent empirical study by Vaze et al. \cite{vaze2022openset} shows that a good closed-set classifier can be competitive with or even superior to previous complex methods. These findings align with our results on UniDA under the foundation models.

\section{Problem formulation}
In UniDA, we are provided with a source domain dataset $\mathcal{D}^s = \{(\mathbf{x}^s_i,y^s_i)\}_{i=1}^{n_s}$ consisting of $n_s$ samples, where the $i$-th sample $\mathbf{x}^s_i\in \mathbb{R}^{d}$ is a $d$ dimensional vector and $y^s_i \in \mathcal{Y}^s$ is the associated label. Additionally, we have a target domain dataset $\mathcal{D}^t = \{(\mathbf{x}^t_i)\}_{i=1}^{n_t}$, which contains $n_t$ unlabeled samples from the same $d$-dimensional space. Samples in the source and target domains are drawn from their respective distributions, $\mathcal{D}^s \sim \pi_s(X^s, Y^s)$ and $\mathcal{D}^t \sim \pi_t(X^t, Y^t)$.
We represent the collection of labels in the source domain as $\mathcal{Y}^s$ and in the target domain as $\mathcal{Y}^t$. Let $\mathcal{Y}^{st} = \mathcal{Y}^s\cap \mathcal{Y}^t$ be the domain-shared label set and $\mathcal{Y}^{t/s} = \mathcal{Y}^t\setminus \mathcal{Y}^s$ be the target-private label set. Similarly, $\mathcal{Y}^{s/t}$ is the set of source-private labels. In UniDA, we make no assumptions about $\mathcal{Y}^t$. Hence, $\mathcal{Y}^{st}$ and $\mathcal{Y}^{t/s}$ are also unknown. For convenience, we refer to target samples belonging to $\mathcal{Y}^{st}$ (known classes) as in-class samples $\mathcal{D}^t_{in}$ and those belonging to $\mathcal{Y}^{t/s}$ (unknown classes) as out-class samples $\mathcal{D}^t_{out}$.

The learning task of UniDA can be converted as two subtasks of in-class discrimination and out-class detection. Such objectives could be implemented by a unified framework as: (1) learning a scoring function $s: \mathbb{R}^d \rightarrow \mathbb{R}$ for out-class detection and (2) learning a classifier $f: \mathbb{R}^d \rightarrow \mathbb{R}^{|\mathcal{Y}^s|}$ for in-class discrimination. The scoring function $s$ assigns a score to each sample, which reflects the uncertainty level regarding it being an out-class sample. A higher score indicates a higher likelihood of belonging to the in-class category. UniDA methods require a threshold value for the scoring function $s$ to distinguish between out-class and in-class samples. This threshold can either be learned automatically or set manually.

Typically, the learning classifier $f=h\circ\phi$ comprises a feature extractor $\phi$ and a classifier head $h$. Prior research in UniDA primarily focuses on fine-tuning $\phi$ using ImageNet pre-trained backbones. In our study, we aim to explore the training of $f$ and the scoring function $s$ using foundation models such as CLIP backbones.

\section{Empirical analysis of UniDA methods with foundation models}
\subsection{UniDA methods review}\label{section:methods-review}
\begin{table}[t]
\renewcommand{\arraystretch}{1.0}
\centering
\begin{adjustbox}{width=1.0\textwidth,center}
\begin{tabular}{c | c c c c c c}
 \toprule
 Method type & Methods & Source & Target & Classifier & Scoring rule & Threshold value \\
 \midrule
 Baseline & Source Only (SO) & \ding{51} & \ding{55} & softmax & negative entropy & $-\log({|\mathcal{Y}^s|}) / 2$ \\ \hline
\multirow{3}{*}{UniDA SOTAs}    & DANCE \cite{saito2020universal} & \ding{51}  & \ding{51} &  softmax & negative entropy & $-\log({|\mathcal{Y}^s|}) / 2$\\
                                  & OVANet \cite{saito2021ovanet} & \ding{51} & \ding{51} & softmax & binary softmax prob. & 1/2\\
                                  & UniOT \cite{chang2022unified} & \ding{51} & \ding{51} &  OT &  maximum OT mass& $1 / (n^t + T)$\\
\hline
\multirow{3}{*}{CLIP adaptations}       & WiSE-FT \cite{wortsman2022robust} & \ding{51}  & \ding{55} & softmax& negative entropy & $-\log({|\mathcal{Y}^s|}) / 2$\\
                                    & CLIP cross-model \cite{lin2023multimodality} & \ding{51} & \ding{55} & softmax & negative entropy & $-\log({|\mathcal{Y}^s|}) / 2$ \\
                                    & CLIP zero-shot \cite{radford2021learning} & \ding{55} & \ding{55} & NN &  maximum logit & -\\ \hline
  Ours                              & CLIP distillation & \ding{55} & \ding{51} & softmax & negative entropy & $-\log({|\mathcal{Y}^s|}) / 2$\\
\bottomrule
\end{tabular}
\end{adjustbox}
\caption{A brief introduction of different methods. Previous approaches are categorized into three groups: SO, the baseline method that trains models solely on source data; DANCE \cite{saito2020universal}, OVANet \cite{saito2021ovanet}, and UniOT \cite{chang2022unified}, state-of-the-art methods designed explicitly for the UniDA task; and WiSE-FT \cite{wortsman2022robust}, CLIP cross-model \cite{lin2023multimodality}, and CLIP zero-shot \cite{radford2021learning}, three SOTA methods for CLIP model adaptation. In this paper, we introduce CLIP distillation, as detailed in Section \ref{section:proposed-method}. Note that our method uses the source data for confidence calibration, not for training.}
\label{table:methods-list}
\end{table}

\begin{table}[t]
    \renewcommand{\arraystretch}{1.0}
    \centering
    \begin{adjustbox}{width=1.0\textwidth,center}
    \begin{tabular}{c | c c c | c c c | c c c | c c c}
    \toprule
       \multirow{2}{*}{Methods}  &  \multicolumn{3}{|c|}{ImageNet-pretrained} & \multicolumn{3}{|c|}{Random initialization} & \multicolumn{3}{|c|}{DINOv2-pretrained} &  \multicolumn{3}{|c}{CLIP-pretrained}\\ \cline{2-13}
        & H-score & H$^3$-score & UCR & H-score & H$^3$-score & UCR & H-score & H$^3$-score & UCR & H-score & H$^3$-score & UCR \\ \midrule
        \multicolumn{13}{c}{Fine-tuning backbone} \\ \midrule
        SO &  58.16  &  63.73 & 52.31  &  8.01 & 1.45  & 6.78  & 0.54  & 0.46  &  7.16  &  2.15 &  1.95  &  8.46  \\
        DANCE\cite{saito2020universal} &  42.36 & 49.17 &  28.36  & 0.82  &  1.02 & 5.81  & 0.42  & 0.44  & 5.94  & 0.46  & 0.58  &  5.98  \\
        OVANet\cite{saito2021ovanet} & 38.27  & 45.60  & 53.63  &  1.55 & 1.23  & 7.36  & 6.65  & 0.82  & 5.51  & 1.06  & 1.08  & 4.40   \\
        UniOT\cite{chang2022unified} & 71.23  & 70.93  & 65.52  & 11.95  & 2.17  & 9.44  &  7.56 & 2.13  & 5.98  & 15.02  & 2.67  & 8.42   \\ \midrule
        \multicolumn{13}{c}{Freeze backbone} \\ \midrule
        SO     &  56.69  & 62.19  & 52.90  &  14.16 & 1.73  & 5.09  & 57.63  & 65.16  & 53.12 &  70.30  & 73.58  &  67.28   \\
        DANCE\cite{saito2020universal}  &  42.59  & 50.07  & 28.12  &  10.08 & 1.67  & 6.40  &  44.79 & 53.58  &  34.53  & 67.79  &  71.73  &  54.17  \\
        OVANet\cite{saito2021ovanet} &  56.36  & 62.75  & 67.77  & 6.55 & 1.58  & 4.30 & 57.91  & 65.40  & 48.86   & 56.36  &  62.75  & 67.77   \\
        UniOT\cite{chang2022unified}  &  68.26  & 67.16  & 62.25  & 11.24  & 1.39  &  7.59 & 62.73  & 67.12  & 52.27  & 75.87  & 74.81   & 67.58   \\
        \bottomrule
    \end{tabular}
    \end{adjustbox}
    \caption{Comparison results using ViT-B \cite{vaswani2017attention} backbone with various pre-trained models and fine-tuning modes. Results are conducted on the VisDA dataset in the open-partial UniDA setting.}
    \label{tab:fine-tuing-or-not}
\end{table}
We briefly review some representative state-of-the-art (SOTA) methods for comparison: DANCE \cite{saito2020universal}, OVANet \cite{saito2021ovanet}, and UniOT \cite{chang2022unified}. Notably, certain other methods are not included in the comparison table due to their inferior performance when compared to these approaches. For a concise overview of these methods, see Table \ref{table:methods-list}.

\noindent \textbf{Source Only (SO, baseline)}. The Source Only (SO) method involves standard cross-entropy loss training on the source data alone. In inference, the softmax classifier $f$ is employed for predictions, and the scoring function $s$ is constructed based on the entropy of the softmax output probabilities, following the approach used in DANCE.

\noindent \textbf{DANCE}\cite{saito2020universal}. This approach not only utilizes standard training on the source data but also introduces a self-supervised loss for target feature clustering and an entropy separation loss to either align target features with the source domain or classify them as unknown classes. Following the training, a softmax classifier $f$ is learned based on the similarities between the target features and the source prototypes. The scoring function $s$ is calculated as the negative entropy of the softmax output. The threshold for out-class detection is set as $-\log({|\mathcal{Y}^s|}) / 2$.

\noindent \textbf{OVANet}\cite{saito2021ovanet}. This approach introduces a one-vs-all network for tackling the UniDA task, consisting of a standard source classifier $f$ and $|\mathcal{Y}^s|$ binary softmax classifiers. Throughout training, each binary classifier is responsible for distinguishing source samples in its respective class from those in other classes. Once trained, the scoring function $s$ is constructed based on the output probability of a chosen binary classifier. The threshold for out-class detection is set at 0.5.

\noindent \textbf{UniOT}\cite{chang2022unified}. Similar to DANCE, this approach also employs self-supervised learning to achieve target clustering and alignment. It does so by constructing source prototypes and numerous target prototypes. However, what sets it apart is its use of Optimal Transport (OT) to select confident target samples and prototypes to achieve its objective. Finally, the classifier $f$ makes predictions based on the outcomes of the optimal transport process between target features and source prototypes. The scoring function $s$ is created based on the maximum OT probability of target samples. The threshold ($1/(n^t + T)$) for identifying out-class samples is dynamically adjusted according to $T$, which is task-specific.

\subsection{Key observations and suggestions}\label{sec:clip-is-the-best}
\begin{table}[t]
\renewcommand{\arraystretch}{1.0}
\centering
\begin{adjustbox}{width=1.0\textwidth,center}
\begin{tabular}{c | c c c c |c| c c c c |c| c c c c|c}
\toprule
\multirow{2}{*}{Methods} & \multicolumn{5}{|c|}{Resnet50} & \multicolumn{5}{|c|}{DINOv2} & \multicolumn{5}{|c}{CLIP}\\ \cline{2-16}
                         & Office & OH & VD & DN & Avg & Office & OH & VD & DN & Avg & Office & OH & VD & DN & Avg \\
\midrule
\multicolumn{16}{c}{H-score} \\
\midrule
SO	&	66.34	&	54.39	&	30.33	&	39.2	&	47.56	&	89.6	&	82.77	&	57.53	&	68.38	&	74.57	&	91.98	&	84.52	&	69.85	&	61.49	&	76.96	\\
DANCE\cite{saito2020universal}	&	80.32	&	39.06	&	2.78	&	26.91	&	37.27	&	\textbf{91.93}	&	84.38	&	53.89	&	68.74	&	74.73	&	\textbf{94.7}	&	89.01	&	71.9	&	60.53	&	79.03	\\
OVANet\cite{saito2021ovanet}	&	83.33	&	71.68	&	44.57	&	49.57	&	62.29	&	86.51	&	76.83	&	\textbf{58.03}	&	55.76	&	69.28	&	93.36	&	85.42	&	59.47	&	70.7	&	77.24	\\
UniOT\cite{chang2022unified}	&	\textbf{84.37}	&	\textbf{75.97}	&	\textbf{54.48}	&	\textbf{50.88}	&	\textbf{66.42}	&	89.16	&	\textbf{87.54}	&	56.6	&	\textbf{69.86}	&	\textbf{75.79}	&	92.32	&	\textbf{89.45}	&	\textbf{79.1}	&	\textbf{71.42}	&	\textbf{83.07}	\\
\midrule
\multicolumn{16}{c}{H$^3$-score} \\
\midrule
SO	&	65.55	&	56.87	&	34.92	&	42.16	&	49.87	&	88.74	&	82.81	&	65.03	&	70.9	&	\textbf{76.87}	&	89.95	&	82.7	&	74.24	&	64.65	&	77.88	\\
DANCE\cite{saito2020universal}	&	71.57	&	44.0	&	4.04	&	31.18	&	37.7	&	\textbf{90.3}	&	83.88	&	61.84	&	\textbf{71.14}	&	76.79	&	\textbf{91.64}	&	85.6	&	75.77	&	63.94	&	79.24	\\
OVANet\cite{saito2021ovanet}	&	76.8	&	67.96	&	45.91	&	\textbf{50.34}	&	60.25	&	86.64	&	78.65	&	\textbf{65.45}	&	60.26	&	72.75	&	90.8	&	83.33	&	66.07	&	\textbf{71.1}	&	77.82	\\
UniOT\cite{chang2022unified}	&	\textbf{77.33}	&	\textbf{70.49}	&	\textbf{46.01}	&	47.91	&	\textbf{60.43}	&	87.68	&	\textbf{85.5}	&	61.94	&	70.43	&	76.39	&	89.07	&	\textbf{87.09}	&	\textbf{77.69}	&	69.9	&	\textbf{80.94}	\\
\midrule
\multicolumn{16}{c}{UCR} \\
\midrule
SO	&	81.21	&	65.18	&	24.59	&	31.04	&	50.5	&	\textbf{91.2}	&	\textbf{84.46}	&	\textbf{50.36}	&	61.45	&	\textbf{71.87}	&	93.98	&	86.89	&	63.46	&	63.19	&	76.88	\\
DANCE\cite{saito2020universal}	&	84.47	&	69.34	&	\textbf{44.13}	&	32.71	&	57.66	&	87.32	&	83.32	&	41.33	&	\textbf{63.52}	&	68.87	&	95.17	&	\textbf{90.33}	&	57.78	&	\textbf{64.88}	&	77.04	\\
OVANet\cite{saito2021ovanet}	&	81.38	&	67.83	&	36.26	&	34.43	&	54.97	&	89.96	&	82.03	&	46.48	&	58.6	&	69.27	&	\textbf{95.36}	&	88.18	&	68.57	&	64.3	&	\textbf{79.1}	\\
UniOT\cite{chang2022unified}	&	\textbf{84.74}	&	\textbf{73.65}	&	41.29	&	\textbf{34.52}	&	\textbf{58.55}	&	86.63	&	84.41	&	44.15	&	57.85	&	68.26	&	90.62	&	88.85	&	\textbf{72.22}	&	62.88	&	78.64	\\
\bottomrule
\end{tabular}
\end{adjustbox}
\caption{Comparison results between the baseline (Source Only (SO)) and SOTAs using different backbones in the open-partial UniDA setting.}
\label{table:results-resnet-clip-dino}
\end{table}

\begin{table}[t]
\renewcommand{\arraystretch}{1.0}
\centering
\begin{adjustbox}{width=1.0\textwidth,center}
\begin{tabular}{c | c c c c | c c c c | c c c c | c c c c| c}
\toprule
\multirow{2}{*}{Methods} & \multicolumn{4}{|c|}{Office} & \multicolumn{4}{|c|}{OfficeHome} & \multicolumn{4}{|c|}{VisDA} & \multicolumn{4}{|c|}{DomainNet} & \multirow{2}{*}{Avg}\\ \cline{2-17}
                         & (10/10) & (10/0) & (31/0) & (10/21) & (10/5) & (15/0) & (65/0) & (25/40) & (6/3) & (6/0) & (12/0) & (6/6) & (150/50) & (150/0) & (345/0) & (150/195) & \\
\midrule
\multicolumn{18}{c}{H-score} \\
\midrule
SO	&	89.6	&	92.29	&	\textbf{83.99}	&	90.5	&	82.77	&	81.76	&	66.72	&	64.41	&	57.53	&	64.49	&	42.45	&	38.06	&	68.38	&	70.36	&	51.66	&	50.53	&	68.47	\\
DANCE\cite{saito2020universal}	&	\textbf{91.93}	&	\textbf{94.98}	&	81.67	&	80.32	&	84.38	&	81.93	&	64.28	&	57.03	&	53.89	&	65.26	&	34.87	&	26.49	&	68.74	&	70.51	&	51.59	&	49.3	&	66.07	\\
OVANet\cite{saito2021ovanet}	&	86.51	&	88.7	&	82.17	&	\textbf{92.02}	&	76.83	&	76.2	&	\textbf{70.6}	&	\textbf{71.29}	&	\textbf{58.03}	&	62.44	&	\textbf{56.91}	&	\textbf{61.51}	&	55.76	&	57.49	&	\textbf{58.92}	&	\textbf{58.51}	&	\textbf{69.62}	\\
UniOT\cite{chang2022unified}	&	89.16	&	94.52	&	65.44	&	41.04	&	\textbf{87.54}	&	\textbf{85.56}	&	55.81	&	38.55	&	56.6	&	\textbf{71.56}	&	39.31	&	29.62	&	\textbf{69.86}	&	\textbf{72.64}	&	54.0	&	45.08	&	62.27	\\
\midrule
\multicolumn{18}{c}{H$^3$-score} \\
\midrule
SO	&	88.74	&	90.73	&	\textbf{83.99}	&	90.5	&	82.81	&	82.15	&	66.72	&	64.41	&	65.03	&	66.31	&	42.45	&	38.06	&	70.9	&	72.2	&	51.66	&	50.53	&	69.2	\\
DANCE\cite{saito2020universal}	&	\textbf{90.3}	&	\textbf{92.48}	&	81.67	&	80.32	&	83.88	&	82.22	&	64.28	&	57.03	&	61.84	&	66.85	&	34.87	&	26.49	&	\textbf{71.14}	&	\textbf{72.3}	&	51.59	&	49.3	&	66.66	\\
OVANet\cite{saito2021ovanet}	&	86.64	&	88.38	&	82.17	&	\textbf{92.02}	&	78.65	&	78.24	&	\textbf{70.6}	&	\textbf{71.29}	&	\textbf{65.45}	&	64.83	&	\textbf{56.91}	&	\textbf{61.51}	&	60.26	&	61.67	&	\textbf{58.92}	&	\textbf{58.51}	&	\textbf{71.0}	\\
UniOT\cite{chang2022unified}	&	87.68	&	91.9	&	65.44	&	41.04	&	\textbf{85.5}	&	\textbf{84.11}	&	55.81	&	38.55	&	61.94	&	\textbf{68.67}	&	39.31	&	29.62	&	70.43	&	72.17	&	54.0	&	45.08	&	61.95	\\
\midrule
\multicolumn{18}{c}{UCR} \\
\midrule
SO	&	\textbf{91.2}	&	93.93	&	90.11	&	94.59	&	\textbf{84.46}	&	\textbf{83.82}	&	81.9	&	81.84	&	\textbf{50.36}	&	59.14	&	66.98	&	66.93	&	61.45	&	64.0	&	68.94	&	68.74	&	\textbf{75.52}	\\
DANCE\cite{saito2020universal}	&	87.32	&	94.24	&	86.97	&	85.2	&	83.32	&	82.04	&	80.1	&	75.76	&	41.33	&	53.03	&	52.94	&	44.51	&	\textbf{63.52}	&	\textbf{65.85}	&	\textbf{69.73}	&	68.6	&	70.9	\\
OVANet\cite{saito2021ovanet}	&	89.96	&	92.53	&	90.1	&	\textbf{94.64}	&	82.03	&	81.37	&	81.86	&	\textbf{81.84}	&	46.48	&	53.6	&	\textbf{67.0}	&	\textbf{66.98}	&	58.6	&	61.29	&	68.97	&	\textbf{68.74}	&	74.12	\\
UniOT\cite{chang2022unified}	&	86.63	&	\textbf{95.51}	&	\textbf{91.02}	&	59.84	&	84.41	&	82.5	&	\textbf{82.93}	&	57.78	&	44.15	&	\textbf{60.66}	&	64.22	&	40.67	&	57.85	&	63.61	&	69.08	&	60.38	&	68.83	\\
\bottomrule
\end{tabular}
\end{adjustbox}
\caption{Comparison results between the baseline (Source Only (SO)) and SOTAs using DINOv2 backbone in four UniDA settings (open-partial, open, closed, partial).}
\label{table:results-dino}
\end{table}

\noindent \textbf{Fine-tuning backbone or not?}
Table \ref{tab:fine-tuing-or-not} presents a performance comparison among three pre-trained models: ImageNet-pretrained, DINOv2-pretrained, and CLIP-pretrained. We also compare these models to training from scratch, where the backbone is initialized randomly. It is interesting to note that, in contrast to fine-tuning from the ImageNet pre-trained model, fine-tuning from foundation models (DINOv2 or CLIP) often yields significantly poorer results and, in some cases, performs even worse than training from scratch. While keeping the backbones frozen, results of the baseline method (SO) based on foundation models exhibit improvement compared to that based on the ImageNet pre-trained model. This improvement is particularly significant when using CLIP models.
Building upon this observation, we maintain a frozen backbone and only update parameters in other modules when using foundation models.

\noindent \textbf{Why using foundation models?}
Previous UniDA studies have only verified their results using the ImageNet pre-trained Resnet50 model. To demonstrate how these methods perform when using foundation models as their backbones, we present the comparison in Table \ref{table:results-resnet-clip-dino}. It is evident that when employing foundation models as backbones, all methods exhibit a significant improvement in performance. It indicates that the high-level features learned in foundation models are more robust than those in ImageNet pre-trained models.

\noindent \textbf{Which learning algorithm is best?}
Table \ref{table:results-dino} presents a comparative analysis of the baseline and state-of-the-art (SOTA) methods using the DINOv2 foundation model as the backbone. To provide a comprehensive evaluation, we report results across four distinct UniDA scenarios: open-partial, open, closed, and partial, spanning four different UniDA benchmarks, namely, Office, OfficeHome, VisDA, and DomainNet. In terms of H-score/H$^3$-score metrics, UniOT excels in the open-partial and open settings, while the OVANet outperforms other methods in the closed and partial settings. It is noteworthy that when considering the UCR metric, all methods demonstrate similar performance across all tasks. Overall, when averaging over all tasks and all metrics, no single method consistently outperforms the others. This indicates that existing UniDA methods generally fail to improve over the baseline.

\noindent \textbf{DINOv2 or CLIP?}
We selected two of the most powerful visual foundation models, ViT-L/14@336px from CLIP \cite{radford2021learning} and DINOv2 (dinov2\_vitl14) \cite{oquab2023dinov2}, for comparison. The results presented in Tables \ref{table:results-resnet-clip-dino}, \ref{table:results-dino}, and \ref{table:results-all-clip} demonstrate that CLIP models exhibit greater effectiveness than DINOv2 models in UniDA tasks. This observation has motivated us to develop a UniDA method using CLIP foundation models.

In conclusion, we have demonstrated that a powerful backbone is crucial for UniDA tasks. Our findings emphasize the need for further research in the UniDA field, particularly when utilizing these more powerful foundation models.

\begin{figure}
    \centering
    \includegraphics[width=0.8\linewidth]{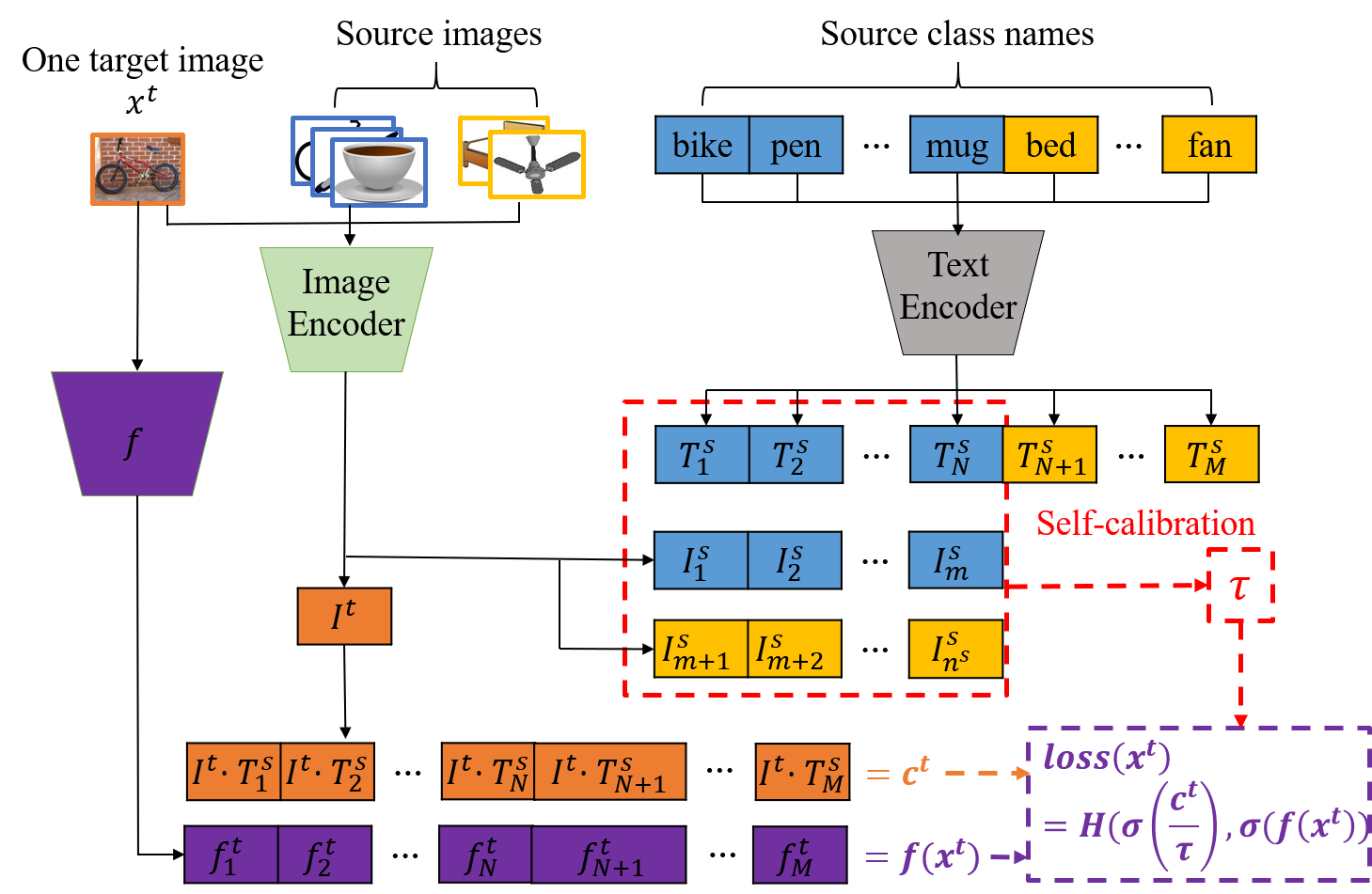}
    \caption{Training pipeline of \textit{CLIP distillation} for UniDA.}
    \label{fig:framework}
\end{figure}
\section{Proposed method}\label{section:proposed-method}
Based on the above observations, we are motivated to adapt from the most powerful CLIP models and propose \textit{CLIP distillation} method. Let $f: \mathbb{R}^d\rightarrow \mathbb{R}^{|\mathcal{Y}^s|}$ be the classifier before softmax layer and $\sigma$ be the softmax function, then the loss of the CLIP distillation to each target example $\mathbf{x}^t$ is:
\begin{equation}
    \text{loss}(\mathbf{x}^t) = H(\sigma(\mathbf{c}^t/\tau), \sigma(f(\mathbf{x}^t))),
\end{equation}
where $\mathbf{c}^t$ is the output logit of $\mathbf{x}^t$ in the CLIP zero-shot model, $H$ is the cross-entropy, and $\tau$ is the scaling temperature, which is automatically learned using the proposed self-calibration method (\ref{sec:self-calibration}). The training framework for \textit{CLIP distillation} is illustrated in Figure \ref{fig:framework}.
\subsection{Motivations}
\noindent \textbf{Why distillation?}
We employ distillation \cite{Hinton2015distillation} to address UniDA using the CLIP model for two main reasons. First, distillation provides a straightforward means to learn from the powerful CLIP model without updating its parameters, thus preventing the risk of destabilizing the CLIP model. Second, given that the CLIP model already yields promising results on the closed-set out-of-distribution (OOD) data \cite{radford2021learning}, distillation from target data resembles a self-training technique, which is theoretically well-grounded \cite{wei2021theoretical}.

\begin{figure}
    \centering
     \begin{subfigure}[b]{0.45\textwidth}
         \centering
         \includegraphics[width=\textwidth]{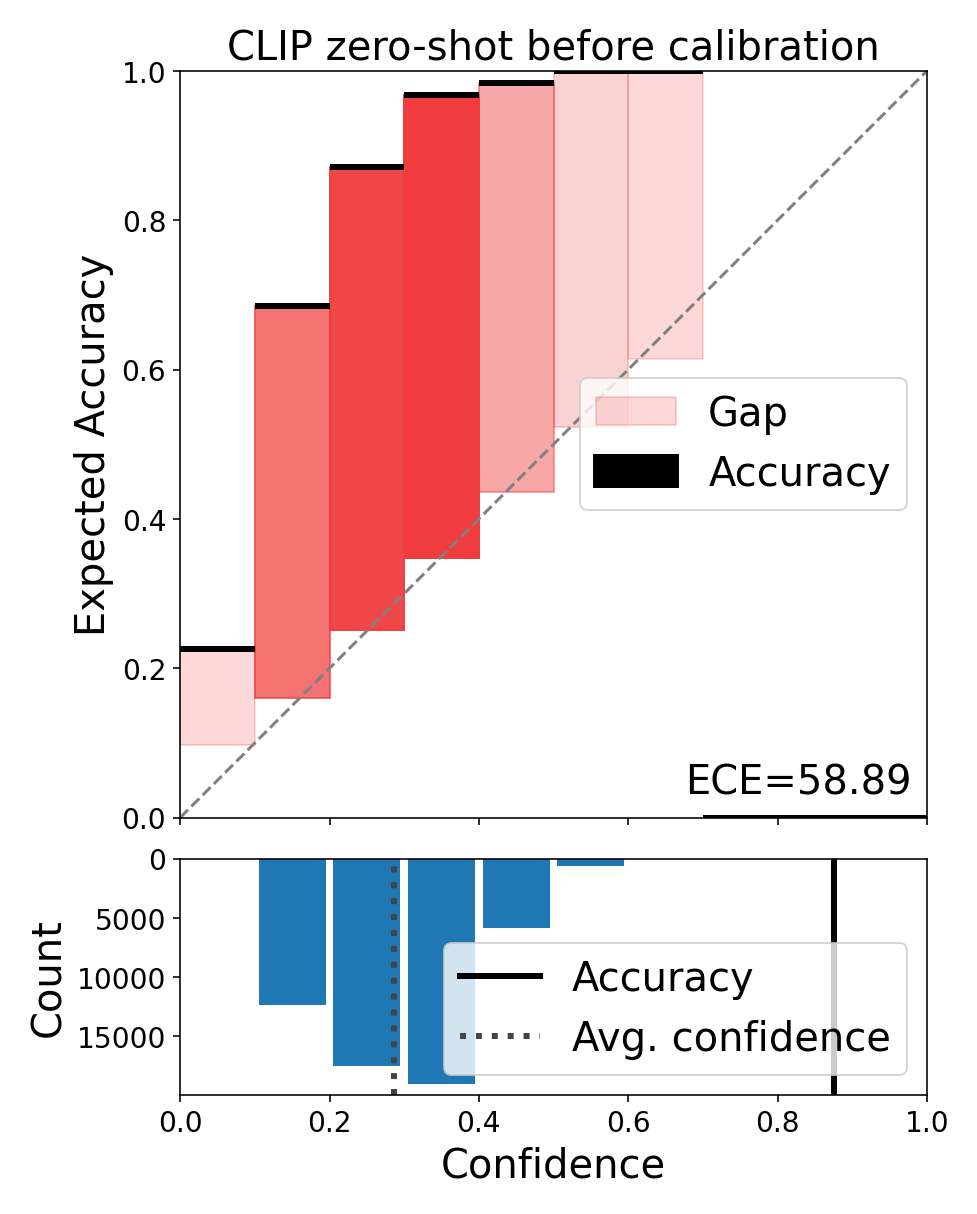}
         \label{fig:CLIP-zeroshot-without-calibration}
     \end{subfigure}
    \centering
     \begin{subfigure}[b]{0.45\textwidth}
         \centering
         \includegraphics[width=\textwidth]{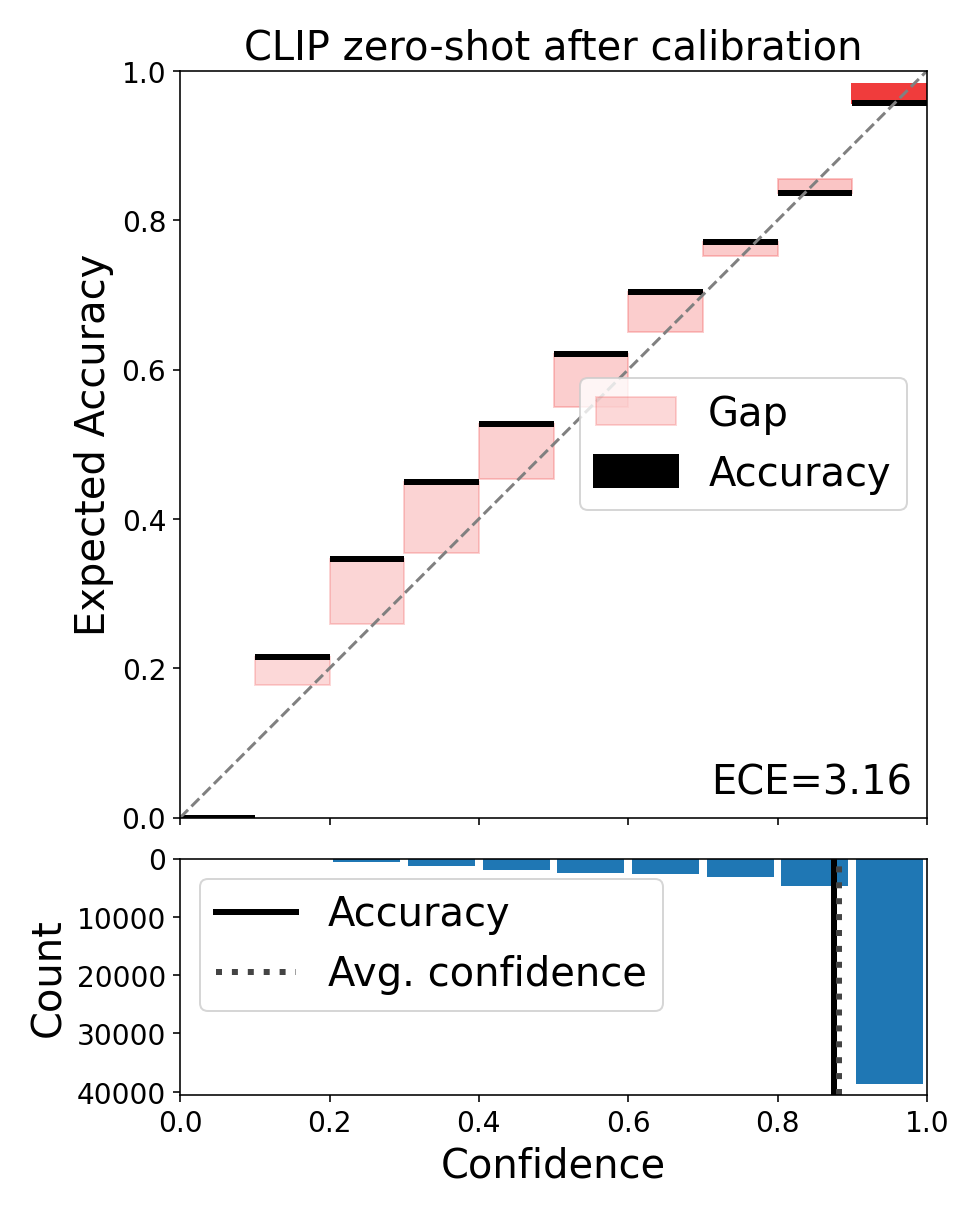}
         \label{fig:CLIP-zeroshot-with-calibration}
     \end{subfigure}
    \caption{Reliability diagrams(top) and confidence histograms (bottom) for CLIP zero-shot model before and after calibration on VisDA dataset.}
    \label{fig:calibration}
\end{figure}

\noindent \textbf{Why scaling the logits?}
In UniDA tasks, however, the objective is not solely closed-set classification but also demands effective out-class detection. We argue that a well-calibrated model plays a pivotal role in achieving this goal. To gain a clearer perspective on our argument, we illustrate two reliability diagrams \cite{degroot1983comparison} comparing the CLIP zero-shot model before and after calibration in Figure \ref{fig:calibration}. As depicted in the figure, without logit scaling, the CLIP zero-shot method tends to classify most samples with low confidence, even if they are classified correctly. This would readily lead to misidentifying the majority of in-class samples as out-class ones, causing a decrease in in-class classification performance. After calibration through temperature scaling, significantly improved confidence estimates can be observed, resulting in a better trustworth prediction system. Therefore, we scale the logits to ensure the model's proper calibration and enhance its performance in both out-class detection and in-class discrimination \cite{guo2017calibration}.

\subsection{Learning temperature scaling by source confidence calibration}\label{sec:self-calibration}
Confidence calibration by temperature scaling faces a challenge for UniDA tasks since we do not have prior knowledge about the target categories. To address this challenge, we propose to learn using the source data. We evenly divide the source data into two parts by class. The first part of the samples is treated as in-class samples for IID calibration, while the second part of the samples is treated as out-class samples for OOD calibration.

\noindent \textbf{IID calibration.}
Given a ground truth joint distribution $\pi_{in}(X,Y)=\pi_{in}(Y|X)\pi_{in}(X)$, the expected calibration error (ECE) for a prediction model is defined as
\begin{equation}
    \mathbb{E}_{\hat{P}}[|\mathbb{P}(\hat{Y}=Y|\hat{P}=p)-p|],
\end{equation}
where $\hat{Y}$ is a class prediction and $\hat{P}$ is its associated confidence, i.e. probability of correctness. ECE could be approximated by partitioning predictions into $K$ equally-spaced bins (similar to the reliability diagrams, see Figure \ref{fig:calibration}) and taking a weight average of the bins' accuracy/confidence difference \cite{naeini2015obtaining}. Specifically,
\begin{equation}
    \text{ECE}_{in} = \sum_{k=1}^K \frac{|B_k|}{n_{in}}|\text{acc}(B_k)-\text{conf}(B_k)|,
\end{equation}
where $n_{in}$ is the number of in-class samples.

\noindent \textbf{OOD calibration.}
For out-class samples $\{x_i\}_{i=1}^{n_{out}}$, which do not belong to any specific predicted category, our objective is to maintain a uniform distribution of their output class probabilities:
\begin{equation}
     \text{ECE}_{out} =  \frac{1}{n_{out}} \sum_{i=1}^{n_{out}}|\text{conf}(x_i)-\frac{1}{N}|,
\end{equation}
where $N$ is the number of in-class categories.

\noindent \textbf{Negative log likelihood.}
As a standard measure of probabilistic model's quality \cite{hastie2009elements}, netagive log likelihood (NLL) is widely used in the context of deep learning, which is also known as the cross entropy loss. Given the known ground truth of in-class samples and a prediction probabilistic model $\hat{\pi}_{in}(Y|X)$, NLL is defined as:
\begin{equation}
    \text{NLL}_{in} = -\sum_{i=1}^{n_{in}} \log(\hat{\pi}_{in}(y_i|x_i))
\end{equation}
$\text{NLL}_{in}$ is minimized if and only if $\hat{\pi}_{in}(Y|X)$ recovers the ground truth conditional distribution $\pi_{in}(Y|X)$.

Our overall objective of learning temperature scaling is then written as
\begin{equation}
    \tau_{opt} = \arg \min_{\tau}\text{ECE}_{in} + \text{ECE}_{out} + \text{NLL}_{in}.
\end{equation}

\section{Experiments}
In this section, we conduct a comparative analysis of our method to demonstrate its robustness and effectiveness in tackling UniDA. We employ CLIP as the backbone due to its superior performance, as outlined in Section \ref{sec:clip-is-the-best}.
\subsection{Datasets and experimental setup}
\noindent \textbf{Dataset.} We train the above methods on the standard benchmark datasets for UniDA: Office \cite{office2010adapting}, OfficeHome (OH) \cite{offhome2017deep}, VisDA (VD) \cite{visda2017visda}, and DomainNet (DN) \cite{domainnet2019moment}. Office has 31 categories and three domains: Amazon (A), DSLR (D), and Webcam (W). OfficeHome contains 65 categories and four domains: Art (A), Clipart (C), Product (P), and Real-World (R) images. VisDA is a synthetic-to-real dataset with 12 categories in total. DomainNet is the largest dataset, including 345 categories and six domains, where three domains -- Painting (P), Real (R), and Sketch (S) -- are used in experiments following previous work \cite{saito2021ovanet, chang2022unified}. For each dataset, we further split the total categories into three disjoint parts -- common categories $\mathcal{Y}^{st}$, source private categories $\mathcal{Y}^{s/t}$, and target private categories $\mathcal{Y}^{t/s}$ -- to consist of the source and target domains. For a more comprehensive study, we assign each dataset with four different class splits: open-partial, open, closed, partial, following \cite{saito2020universal}. The different classes splits result in different running tasks, and each split setting, denoted ($|\mathcal{Y}^{st}|/|\mathcal{Y}^{s/t}|$), is shown in each table. The detail information about these four datasets and the class-split settings are presented in Appendix.

\noindent \textbf{Implementation.} For fair comparison between different methods, we implement UniOOD, a code framework to streamline rigorous and reproducible experiments in UniDA. By using the UniOOD framework, all methods are run under the same learning setting. The initial learning rate is set to 0.01 for all new layers and 0.001 for pre-trained backbone if it is fine-tuned and decays using the cosine schedule rule with a warmup of 50 iterations. We use SGD optimizer with momentum 0.9 and the batch size is set to 32 for each domain. The number of training iterations are set to 5000, 10000, or 20000 based on the scale of the training data, which is detailed in Appendix. We report results of the last checkpoint due to the absence of validation data and average them among three random runs. Due to space constraints, we provide the average results for each split setting, while the detailed results for individual tasks can be found in the appendix. Hyperparameters for previous methods follow their official codes.
We do not use any data augmentation during training for fair comparison to different methods, which may be different from previous works.

\begin{table}[t]
\renewcommand{\arraystretch}{1.0}
\centering
\begin{adjustbox}{width=1.0\textwidth,center}
\begin{tabular}{c | c c c c | c c c c | c c c c | c c c c | c}
\toprule
\multirow{2}{*}{Methods} & \multicolumn{4}{|c|}{Office} & \multicolumn{4}{|c|}{OfficeHome} & \multicolumn{4}{|c|}{VisDA} & \multicolumn{4}{|c|}{DomainNet} & \multirow{2}{*}{Avg}\\ \cline{2-17}
                         & (10/10) & (10/0) & (31/0) & (10/21) & (10/5) & (15/0) & (65/0) & (25/40) & (6/3) & (6/0) & (12/0) & (6/6) & (150/50) & (150/0) & (345/0) & (150/195) & \\
\midrule
\multicolumn{18}{c}{H-score} \\
\midrule
SO	&	91.98	&	91.87	&	80.22	&	89.79	&	84.52	&	82.05	&	58.12	&	58.31	&	69.85	&	75.79	&	55.31	&	57.19	&	61.49	&	65.63	&	38.27	&	35.88	&	68.52	\\
DANCE\cite{saito2020universal}	&	\textbf{94.7}	&	96.09	&	75.76	&	66.83	&	89.01	&	83.95	&	55.42	&	46.63	&	71.9	&	74.3	&	58.08	&	49.5	&	60.53	&	65.24	&	37.56	&	30.92	&	66.03	\\
OVANet\cite{saito2021ovanet}	&	93.36	&	91.16	&	74.64	&	87.53	&	85.42	&	80.29	&	64.65	&	65.92	&	59.47	&	39.27	&	43.55	&	42.58	&	70.7	&	72.4	&	57.22	&	55.86	&	67.75	\\
UniOT\cite{chang2022unified}	&	92.32	&	\textbf{96.48}	&	59.95	&	41.31	&	\textbf{89.45}	&	\textbf{86.64}	&	59.27	&	43.6	&	79.1	&	\textbf{83.08}	&	71.62	&	62.03	&	71.42	&	73.21	&	\textbf{63.72}	&	55.18	&	70.52	\\ \hline
WiSE-FT\cite{wortsman2022robust}	&	82.34	&	94.07	&	47.87	&	53.57	&	79.37	&	73.44	&	13.64	&	16.56	&	62.68	&	72.21	&	30.05	&	27.4	&	3.74	&	7.92	&	0.3	&	0.29	&	41.59	\\
CLIP cross-model\cite{lin2023multimodality}	&	93.04	&	93.59	&	83.21	&	92.55	&	86.2	&	84.26	&	62.65	&	63.14	&	77.69	&	81.66	&	62.08	&	67.98	&	61.98	&	67.1	&	36.2	&	34.06	&	71.71	\\ \hline
CLIP distillation ($\tau=1$)	&	0.0	&	0.07	&	0.0	&	0.0	&	0.17	&	1.31	&	0.0	&	0.0	&	0.0	&	0.05	&	0.0	&	0.0	&	0.0	&	0.0	&	0.0	&	0.0	&	0.1	\\
CLIP distillation (Ours)	&	87.46	&	91.84	&	83.34	&	94.32	&	87.37	&	85.37	&	77.76	&	79.52	&	84.73	&	82.24	&	74.03	&	81.83	&	\textbf{73.48}	&	\textbf{74.64}	&	60.16	&	65.89	&	80.25	\\
CLIP distillation (Ours, fixed model)	&	86.74	&	91.89	&	\textbf{83.8}	&	\textbf{94.39}	&	86.4	&	84.77	&	\textbf{80.58}	&	\textbf{80.81}	&	\textbf{84.74}	&	82.26	&	\textbf{76.08}	&	\textbf{83.12}	&	72.37	&	74.22	&	\textbf{74.93}	&	\textbf{75.22}	&	\textbf{82.02}	\\
\midrule
\multicolumn{18}{c}{H$^3$-score} \\
\midrule
SO	&	89.95	&	89.79	&	80.22	&	89.79	&	82.7	&	81.14	&	58.12	&	58.31	&	74.24	&	76.9	&	55.31	&	57.19	&	64.65	&	67.5	&	38.27	&	35.88	&	68.75	\\
DANCE\cite{saito2020universal}	&	\textbf{91.64}	&	92.4	&	75.76	&	66.83	&	85.6	&	82.43	&	55.42	&	46.63	&	75.77	&	75.86	&	58.08	&	49.5	&	63.94	&	67.22	&	37.56	&	30.92	&	65.97	\\
OVANet\cite{saito2021ovanet}	&	90.8	&	89.42	&	74.64	&	87.53	&	83.33	&	79.94	&	64.65	&	65.92	&	66.07	&	47.2	&	43.55	&	42.58	&	71.1	&	72.1	&	57.22	&	55.86	&	68.24	\\
UniOT\cite{chang2022unified}	&	89.07	&	\textbf{93.24}	&	59.95	&	41.31	&	\textbf{87.09}	&	\textbf{85.16}	&	59.27	&	43.6	&	77.69	&	78.17	&	71.62	&	62.03	&	69.9	&	70.83	&	\textbf{63.72}	&	55.18	&	69.24	\\ \hline
WiSE-FT\cite{wortsman2022robust}	&	83.4	&	91.19	&	47.87	&	53.57	&	79.4	&	75.32	&	13.64	&	16.56	&	68.68	&	74.4	&	30.05	&	27.4	&	5.46	&	11.21	&	0.3	&	0.29	&	42.42	\\
CLIP cross-model\cite{lin2023multimodality}	&	90.56	&	90.87	&	83.21	&	92.55	&	83.73	&	82.54	&	62.65	&	63.14	&	79.96	&	80.82	&	62.08	&	67.98	&	65.02	&	68.54	&	36.2	&	34.06	&	71.49	\\ \hline
CLIP distillation ($\tau=1$)	&	0.0	&	0.11	&	0.0	&	0.0	&	0.25	&	1.93	&	0.0	&	0.0	&	0.0	&	0.08	&	0.0	&	0.0	&	0.0	&	0.0	&	0.0	&	0.0	&	0.15	\\
CLIP distillation (Ours)	&	86.9	&	89.74	&	83.34	&	94.32	&	84.39	&	83.18	&	77.76	&	79.52	&	84.8	&	81.2	&	74.03	&	81.83	&	\textbf{73.0}	&	\textbf{73.6}	&	60.16	&	65.89	&	79.6	\\
CLIP distillation (Ours, fixed model)	&	86.45	&	89.77	&	\textbf{83.8}	&	\textbf{94.39}	&	83.73	&	82.74	&	\textbf{80.58}	&	\textbf{80.81}	&	\textbf{84.8}	&	\textbf{81.21}	&	\textbf{76.08}	&	\textbf{83.12}	&	72.08	&	73.18	&	\textbf{74.93}	&	\textbf{75.22}	&	\textbf{81.43}	\\
\midrule
\multicolumn{18}{c}{UCR} \\
\midrule
SO	&	93.98	&	94.95	&	91.44	&	96.99	&	86.89	&	84.53	&	83.55	&	84.44	&	63.46	&	71.17	&	76.66	&	79.51	&	63.19	&	66.01	&	71.26	&	70.78	&	79.93	\\
DANCE\cite{saito2020universal}	&	95.17	&	97.09	&	87.69	&	81.1	&	90.33	&	86.76	&	81.74	&	75.63	&	57.78	&	63.45	&	67.86	&	56.4	&	64.88	&	68.37	&	71.66	&	67.6	&	75.84	\\
OVANet\cite{saito2021ovanet}	&	95.36	&	95.8	&	91.4	&	96.84	&	88.18	&	85.72	&	83.52	&	84.37	&	68.57	&	66.45	&	76.68	&	79.58	&	64.3	&	66.94	&	71.37	&	70.94	&	80.38	\\
UniOT\cite{chang2022unified}	&	90.62	&	97.2	&	92.14	&	55.94	&	88.85	&	85.59	&	85.5	&	63.61	&	72.22	&	78.8	&	84.79	&	74.78	&	62.88	&	67.33	&	73.85	&	67.99	&	77.63	\\ \hline
WiSE-FT\cite{wortsman2022robust}	&	95.27	&	96.33	&	92.3	&	97.57	&	90.77	&	89.28	&	87.28	&	88.44	&	70.83	&	77.88	&	81.45	&	84.43	&	68.72	&	71.66	&	75.74	&	75.77	&	83.98	\\
CLIP cross-model\cite{lin2023multimodality}	&	\textbf{95.38}	&	96.18	&	\textbf{93.24}	&	\textbf{97.58}	&	89.71	&	87.82	&	86.95	&	87.97	&	73.22	&	79.06	&	81.15	&	83.76	&	68.81	&	71.53	&	75.57	&	75.55	&	83.97	\\
CLIP zero-shot\cite{radford2021learning}	&	90.1	&	97.68	&	87.69	&	96.61	&	90.21	&	89.67	&	89.08	&	89.43	&	78.6	&	82.86	&	87.56	&	88.1	&	70.78	&	73.34	&	79.48	&	79.87	&	85.69	\\ \hline
CLIP distillation ($\tau=1$)	&	92.46	&	97.75	&	87.68	&	96.61	&	92.91	&	\textbf{91.71}	&	89.08	&	89.41	&	80.9	&	85.75	&	87.56	&	88.1	&	69.2	&	72.81	&	79.48	&	79.88	&	86.33	\\
CLIP distillation (Ours)	&	93.76	&	\textbf{97.92}	&	87.88	&	96.61	&	92.91	&	91.49	&	\textbf{89.71}	&	\textbf{89.91}	&	\textbf{82.59}	&	\textbf{86.39}	&	\textbf{88.11}	&	\textbf{88.81}	&	73.08	&	74.93	&	\textbf{80.33}	&	\textbf{82.03}	&	\textbf{87.28}	\\
CLIP distillation (Ours, fixed model)	&	93.39	&	97.91	&	87.69	&	96.61	&	\textbf{93.02}	&	\textbf{91.73}	&	89.08	&	89.43	&	82.38	&	86.37	&	87.56	&	88.1	&	\textbf{74.86}	&	\textbf{77.21}	&	79.49	&	79.87	&	87.17	\\
\bottomrule
\end{tabular}
\end{adjustbox}
\caption{Comparison results between existing methods and the proposed method using CLIP backbone in four UniDA settings (open-partial, open, closed, partial).}
\label{table:results-all-clip}
\end{table}

\subsection{Evaluation and discussion}\label{section:previous-metrics}
As Universal Domain Adaptation (UniDA) encompasses a dual objective, it aims to not only reject samples from unknown classes $\mathcal{Y}^{t/s}$ but also accurately classify samples from the correct classes $\mathcal{Y}^{st}$. This makes the evaluation of the UniDA method more complex. There are various evaluation metrics designed to handle the unknown classes $\mathcal{Y}^{t/s}$ in different ways. However, each of these metrics has certain drawbacks, which we discuss in detail for each of them.

\subsubsection{Hard out-class detection criteria}
\noindent \textbf{Average class accuracy}: The initial metric used to evaluate UniDA is the average class accuracy, calculated over a total of $|\mathcal{Y}^{st}|+1$ classes, including all unknown classes $\mathcal{Y}^{t/s}$ grouped together as a superclass \cite{you2019universal}. The drawback of this metric is that it is highly sensitive to the number of shared classes $\mathcal{Y}^{st}$. Having a significant number of shared classes would undeniably render the detection of unknown classes trivial. However, in such scenarios, there is a possibility that the number of out-class samples exceeds the number of in-class samples by multiple folds. One may argue that a simplified weighted accuracy might solve this issue, but we lack prior knowledge to determine the appropriate weights for the different classes.

\noindent \textbf{H-score}: The H-score is later proposed to balance the importance of detecting samples outside the class and classifying in the class samples \cite{fu2020learning}. H-score also includes all unknown classes as a superclass but calculates the harmonic mean of the average classes accuracy on known classes (acc$_{in}$) and the accuracy on the superclass (acc$_{out}$), i.e., H-score = 2$\cdot$acc$_{in}\cdot$acc$_{out}$/(acc$_{in}$+acc$_{out}$). This metric is more reasonable than average class accuracy introduced above, but also has a significant bias to the ratios between the numbers of in-class and out-class samples. Due to the lack of prior knowledge to target data, these ratios may diverse in different tasks. It is usually impossible to handle all tasks of different ratios in order to have a fair evaluation between different methods.

\noindent \textbf{H$^3$-score}: In addition to the dual objective of UniDA, the quality of clustering for target private samples is introduced in \cite{chang2022unified} as an additional objective to facilitate the discovery of target private classes. The H$^3$-score is calculated as 3$/((1/$acc$_{in})$+$(1/$acc$_{out})$+$(1/$NMI$))$, where Normalized Mutual Information (NMI) is the widely used metric for clustering. While H$^3$-score provides a more comprehensive evaluation, incorporating additional NMI into UniDA is beyond the scope of the current study. Furthermore, H$^3$-score faces similar challenges as those encountered with H-score.

It is worth noting that in scenarios where the target data lacks unknown classes, the H-score and H$^3$-score metrics lose their applicability and degenerate into acc$_{in}$.

\subsubsection{Soft out-class detection criteria}\label{section:ucr}
The criterias mentioned above require us to classify a sample as either out-class or in-class, which means that we have to set a threshold for out-class detection. In this paper, we are motivated from the field of open set recognition (OSR)  (Open Set Classification Rate (OSCR) \cite{dhamija2018reducing} and the Detection and Identification Rate (DIR)\cite{Phillips2011}) and introduce a new UniDA evaluation metric, which is threshold- and ratio-free.
However, unlike the OSR task, which assumes the absence of source private classes and the presence of target private classes, UniDA is more flexible and does not impose such strict constraints. Therefore, we adapt these metrics to accommodate various UniDA scenarios, introducing a new metric called Universal Classification Rate (UCR).
\begin{figure}[ht] 
    \centering
    \includegraphics[width=0.5\textwidth]{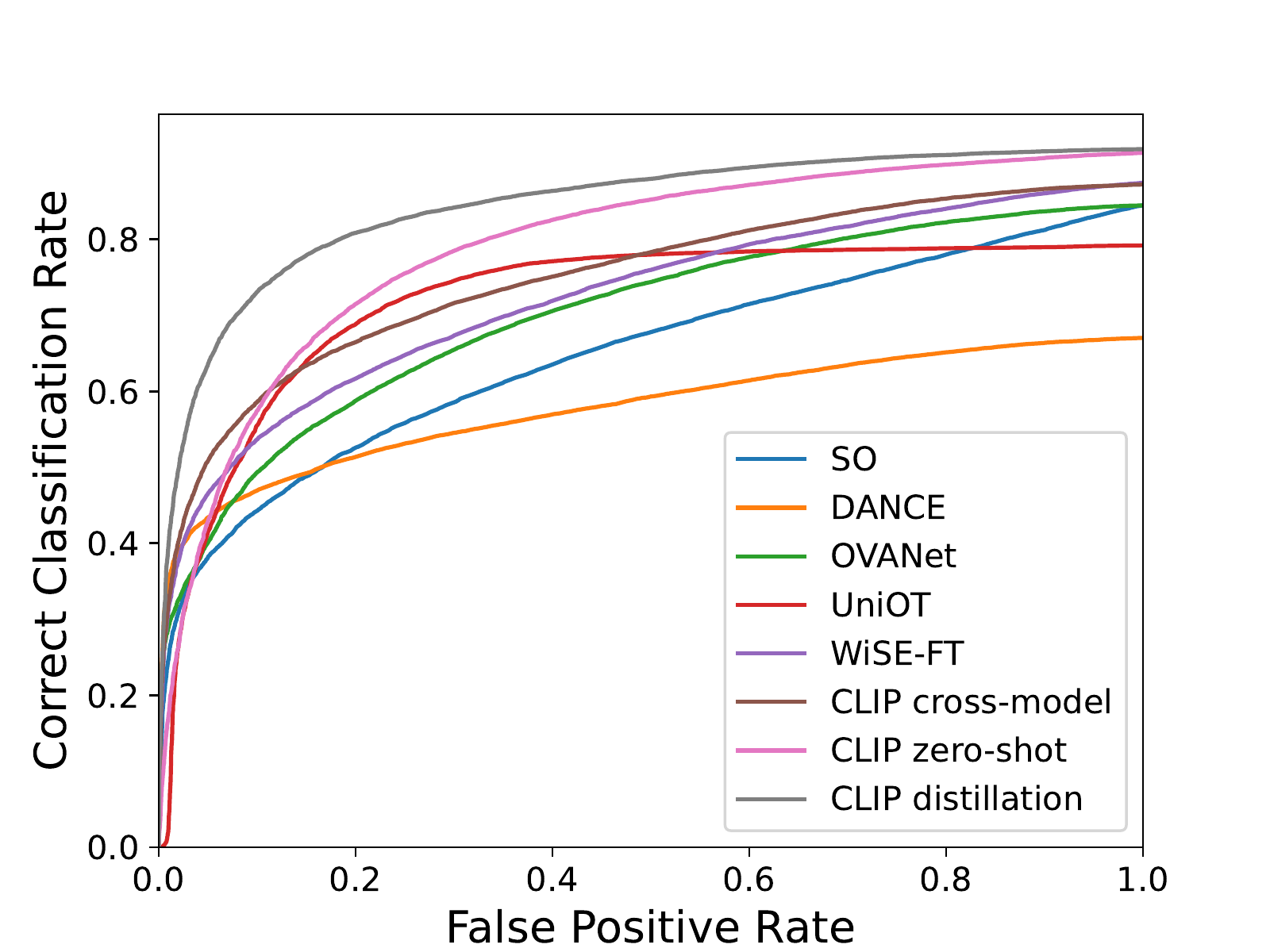}
    \caption{(CCR vs FPR) curve.}
    \label{fig:ucr}
\end{figure}

\noindent \textbf{Universal classification rate (UCR).} To calculate UCR, we compute a pair of Correct Classification Rate (CCR) and False Positive Rate (FPR) by varying the scoring threshold $\theta$. CCR assesses the proportion of correctly classified in-class samples from $\mathcal{D}^t_{in}$, and FPR quantifies the fraction of out-class samples from $\mathcal{D}^t_{out}$ that are incorrectly detected.
\begin{equation}
    \begin{aligned}
        & \text{CCR}(\theta) = \frac{|\{\mathbf{x}|\mathbf{x}\in\mathcal{D}^t_{in} \wedge f(\mathbf{x})=\text{label}(\mathbf{x}) \wedge s(\mathbf{x}) > \theta\}|}{|\mathcal{D}^t_{in}|} \\
        & \text{FPR}(\theta) = \frac{|\{\mathbf{x}|\mathbf{x}\in\mathcal{D}^t_{out} \wedge s(\mathbf{x}) > \theta\}|}{|\mathcal{D}^t_{out}|}.
    \end{aligned}
\end{equation}
Then, the UCR is calculated as
\begin{equation}
    \text{UCR} = \left\{
                            \begin{aligned}
                            & \text{Area Under the (CCR vs FPR) Curve}, \quad \text{if} \ |\mathcal{D}^t_{out}| > 0 \\
                            & \text{CCR($-\infty$)}, \quad \text{if} \ |\mathcal{D}^t_{out}| = 0
                            \end{aligned}
                \right. \\
\end{equation}
where, CCR($-\infty$) is identical to the closed-set classification accuracy on $\mathcal{D}^t_{in}$. Figure \ref{fig:ucr} shows an example illustration to the (CCR vs FPR) curve on VisDA task under the (6/3) setting. The distinction between UCR and AUROC lies in the replacement of the true positive rate (TPR) with the correct classification rate (CCR) in UCR. In contrast to previous evaluation metrics, UCR does not rely on thresholds or ratios, making it an additional criterion that does not consider the threshold effects.

\subsection{Comparison with SOTA UniDA methods}
Table \ref{table:results-all-clip} presents the comparative results between our method and existing state-of-the-art (SOTA) UniDA approaches across three distinct evaluation metrics. The results encompass four distinct UniDA settings, namely open-partial, open, closed, and partial settings, denoted as different class splits represented as ($|\mathcal{Y}^{st}|/|\mathcal{Y}^{s/t}|$) within the table. Regarding the H-score and H$^3$-score metrics, it is evident that our method and the leading state-of-the-art approach are on par with each other in the open-partial and open settings. However, our method exhibits a substantial improvement (>10\%) in six out of eight tasks in the closed and partial settings. In terms of the UCR metric, our method significantly outperforms state-of-the-art UniDA methods in three out of four datasets: OfficeHome, VisDA, and DomainNet, across all four settings. In general, our method exhibits superior robustness across various settings and establishes a new state-of-the-art on UniDA benchmarks, excelling in both the H-score/H$^3$-score metrics and the UCR metric.

\subsection{Comparison with SOTA CLIP-adaptation methods}
Recall that our focus is on developing a UniDA method based on foundation models like CLIP. Therefore, we also provide comparisons with some state-of-the-art (SOTA) adaptation methods that leverage CLIP models, even though they were not originally designed for the UniDA task. These methods include CLIP zero-shot (baseline) \cite{radford2021learning}, WiSE-FT \cite{wortsman2022robust}, and CLIP cross-model \cite{lin2023multimodality}.
WiSE-FT is a new fine-tuning method for improving robustness by ensembling the weights of the zero-shot and fine-tuned models. CLIP cross-model is a recent study introduced by Lin et al. \cite{lin2023multimodality}, which has demonstrated the most remarkable few-shot capability to date by leveraging cross-model information. However, as all these methods can not directly be used for UniDA, we construct a scoring function $s$ following the SO method except for CLIP zero-shot, as illustrated in Table \ref{table:methods-list}.
While these methods have displayed remarkable enhancements in closed-set robustness benchmarks like ImageNet, they frequently exhibit lower performance than the SOTA UniDA methods when evaluated using the H-score/H$^3$-score metric on UniDA benchmarks, as shown in Table \ref{table:results-all-clip}. However, it is noteworthy that all these adaptation methods consistently outperform the SOTA UniDA methods when considering the UCR metric. Our method maintains its position as the most powerful performer in terms of both H-score/H$^3$-score and UCR evaluation metrics.

\subsection{Analysis and ablation study}
\noindent \textbf{Temperature scaling is necessary.} To demonstrate the effectiveness of our temperature scaling, we report the results of the CLIP distillation methods when setting $\tau=1$, as shown in Table \ref{table:results-all-clip}. It's apparent that without temperature scaling, CLIP distillation struggles to distinguish samples between in-class and out-class categories, leading to nearly zero performance on the H-score/H$^3$-score metrics, though its UCR results are marginally lower compared to those with appropriate scaling. This demonstrates the necessity of temperature scaling and the superiority of our self-calibration method.

\noindent \textbf{Distillation helps improve UCR but not H-score.} Comparing the results between CLIP distillation with a model that gets updated and one with a fixed model (Table \ref{table:results-all-clip}), we conclude that while distillation for updating models slightly enhances the UCR, it does not show a consistent improvement in the H-score/H$^3$-score. Nevertheless, distillation provides other advantages when applied to a smaller model.


\begin{table}[t]
\renewcommand{\arraystretch}{1.0}
\centering
\begin{adjustbox}{width=1.0\textwidth,center}
\begin{tabular}{c | c c c c | c c c c | c c c c | c c c c | c}
\toprule
\multirow{2}{*}{Methods} & \multicolumn{4}{|c|}{Office} & \multicolumn{4}{|c|}{OfficeHome} & \multicolumn{4}{|c|}{VisDA} & \multicolumn{4}{|c|}{DomainNet} & \multirow{2}{*}{Avg}\\ \cline{2-17}
                         & (10/10) & (10/0) & (31/0) & (10/21) & (10/5) & (15/0) & (65/0) & (25/40) & (6/3) & (6/0) & (12/0) & (6/6) & (150/50) & (150/0) & (345/0) & (150/195) & \\
\midrule
\multicolumn{18}{c}{H-score} \\
\midrule
w/o IID calibration	&	\textbf{90.31}	&	76.33	&	77.52	&	89.3	&	\textbf{87.61}	&	84.32	&	70.47	&	72.47	&	82.26	&	58.09	&	57.84	&	62.03	&	72.46	&	73.69	&	57.21	&	61.43	&	73.33	\\
w/o NLL calibration	&	88.52	&	73.42	&	74.77	&	86.78	&	87.26	&	83.54	&	64.86	&	67.27	&	78.74	&	54.6	&	54.8	&	58.1	&	71.39	&	72.82	&	55.02	&	58.54	&	70.65	\\
w/o OOD calibration	&	53.55	&	72.84	&	\textbf{86.96}	&	\textbf{96.54}	&	74.54	&	74.21	&	\textbf{83.42}	&	\textbf{86.5}	&	78.86	&	73.11	&	\textbf{78.16}	&	\textbf{86.44}	&	72.97	&	73.83	&	\textbf{61.85}	&	\textbf{68.96}	&	76.42	\\
Ours	&	87.46	&	\textbf{91.84}	&	83.34	&	94.32	&	87.37	&	\textbf{85.37}	&	77.76	&	79.52	&	\textbf{84.73}	&	\textbf{82.24}	&	74.03	&	81.83	&	\textbf{73.48}	&	\textbf{74.64}	&	60.16	&	65.89	&	\textbf{80.25}	\\
\midrule
\multicolumn{18}{c}{H$^3$-score} \\
\midrule
w/o IID calibration	&	\textbf{88.74}	&	79.13	&	77.52	&	89.3	&	\textbf{84.68}	&	82.64	&	70.47	&	72.47	&	83.13	&	63.75	&	57.84	&	62.03	&	72.34	&	72.98	&	57.21	&	61.43	&	73.48	\\
w/o NLL calibration	&	87.5	&	76.96	&	74.77	&	86.78	&	84.49	&	82.15	&	64.86	&	67.27	&	80.7	&	60.91	&	54.8	&	58.1	&	71.61	&	72.41	&	55.02	&	58.54	&	71.05	\\
w/o OOD calibration	&	58.74	&	76.26	&	\textbf{86.96}	&	\textbf{96.54}	&	75.44	&	75.27	&	\textbf{83.42}	&	\textbf{86.5}	&	80.78	&	75.01	&	\textbf{78.16}	&	\textbf{86.44}	&	72.64	&	73.06	&	\textbf{61.85}	&	\textbf{68.96}	&	77.25	\\
Ours	&	86.9	&	\textbf{89.74}	&	83.34	&	94.32	&	84.39	&	\textbf{83.18}	&	77.76	&	79.52	&	\textbf{84.8}	&	\textbf{81.2}	&	74.03	&	81.83	&	\textbf{73.0}	&	\textbf{73.6}	&	60.16	&	65.89	&	\textbf{79.6}	\\
\midrule
\multicolumn{18}{c}{UCR} \\
\midrule
w/o IID calibration	&	93.72	&	97.88	&	87.76	&	96.61	&	93.16	&	91.82	&	89.51	&	89.77	&	82.3	&	86.16	&	87.63	&	88.19	&	73.99	&	75.92	&	80.28	&	81.8	&	87.28	\\
w/o NLL calibration	&	93.65	&	97.88	&	87.74	&	96.61	&	\textbf{93.17}	&	\textbf{91.88}	&	89.43	&	89.75	&	82.18	&	86.15	&	87.61	&	88.16	&	\textbf{74.17}	&	\textbf{76.16}	&	80.24	&	81.7	&	87.28	\\
w/o OOD calibration	&	90.6	&	96.13	&	\textbf{88.18}	&	\textbf{96.67}	&	87.97	&	85.86	&	\textbf{89.99}	&	\textbf{90.15}	&	\textbf{83.41}	&	84.53	&	\textbf{88.56}	&	\textbf{89.67}	&	71.72	&	73.13	&	\textbf{80.39}	&	\textbf{82.24}	&	86.2	\\
Ours	&	\textbf{93.76}	&	\textbf{97.92}	&	87.88	&	96.61	&	92.91	&	91.49	&	89.71	&	89.91	&	82.59	&	\textbf{86.39}	&	88.11	&	88.81	&	73.08	&	74.93	&	80.33	&	82.03	&	\textbf{87.28}	\\
\bottomrule
\end{tabular}
\end{adjustbox}
\caption{Ablation studies.}
\label{table:results-ablation}
\end{table}
\noindent \textbf{Each calibration loss plays a key role.} We conducted ablation studies to assess the significance of each calibration loss component in our method, namely ECE$_{in}$, ECE$_{out}$, and NLL$_{in}$, corresponding to IID calibration, OOD calibration, and NLL calibration, respectively. The results of this analysis are presented in Table \ref{table:results-ablation}. It is evident that the absence of either IID or NLL calibration leads to a substantial decrease in performance in the closed and partial settings. Conversely, the lack of OOD calibration affects the results in the open-partial and open settings. In summary, each calibration loss contributes significantly to the calibration process.

\noindent \textbf{Our method is stable on different CLIP models.}
We present the results of our method when executed on various CLIP models, as outlined in Table \ref{table:results-different-backbones}. The findings reveal that our method exhibits stability when deployed on different backbones, with a modest decrease in performance for smaller models.
\begin{table}[t]
\renewcommand{\arraystretch}{1.0}
\centering
\begin{adjustbox}{width=1.0\textwidth,center}
\begin{tabular}{c | c c c c |c| c c c c |c| c c c c|c}
\toprule
\multirow{2}{*}{} & \multicolumn{5}{|c|}{ViT-B/16} & \multicolumn{5}{|c|}{ViT-L/14} & \multicolumn{5}{|c}{ViT-L/14@336px}\\ \cline{2-16}
                         & Office & OH & VD & DN & Avg & Office & OH & VD & DN & Avg & Office & OH & VD & DN & Avg \\
\midrule
\multirow{6}{*}{open-partial}
& \multicolumn{15}{c}{H-score} \\ \cline{2-16}
&	\text{83.36}	&	\text{84.66}	&	\text{82.11}	&	\text{66.16}	&	\text{79.07}	&	\text{87.19}	&	\text{87.04}	&	\text{84.66}	&	\text{72.54}	&	\text{82.86}	&	\text{87.46}	&	\text{87.37}	&	\text{84.73}	&	\text{73.48}	&	\text{83.26}	\\
\cline{2-16}
& \multicolumn{15}{c}{H$^3$-score} \\ \cline{2-16}
&	\text{81.59}	&	\text{80.22}	&	\text{81.79}	&	\text{67.09}	&	\text{77.67}	&	\text{85.25}	&	\text{84.1}	&	\text{85.04}	&	\text{72.36}	&	\text{81.69}	&	\text{86.9}	&	\text{84.39}	&	\text{84.8}	&	\text{73.0}	&	\text{82.27}	\\
\cline{2-16}
& \multicolumn{15}{c}{UCR} \\ \cline{2-16}
&	\text{81.46}	&	\text{90.4}	&	\text{82.11}	&	\text{64.14}	&	\text{79.53}	&	\text{92.05}	&	\text{92.33}	&	\text{82.58}	&	\text{71.99}	&	\text{84.74}	&	\text{93.76}	&	\text{92.91}	&	\text{82.59}	&	\text{73.08}	&	\text{85.58}	\\

\midrule
\multirow{6}{*}{open}
& \multicolumn{15}{c}{H-score} \\ \cline{2-16}
&	\text{84.22}	&	\text{81.41}	&	\text{84.05}	&	\text{67.81}	&	\text{79.37}	&	\text{90.76}	&	\text{85.14}	&	\text{80.93}	&	\text{73.64}	&	\text{82.62}	&	\text{91.84}	&	\text{85.37}	&	\text{82.24}	&	\text{74.64}	&	\text{83.52}	\\
\cline{2-16}
& \multicolumn{15}{c}{H$^3$-score} \\ \cline{2-16}
&	\text{83.78}	&	\text{78.36}	&	\text{81.43}	&	\text{68.11}	&	\text{77.92}	&	\text{89.4}	&	\text{82.97}	&	\text{80.42}	&	\text{73.1}	&	\text{81.47}	&	\text{89.74}	&	\text{83.18}	&	\text{81.2}	&	\text{73.6}	&	\text{81.93}	\\
\cline{2-16}
& \multicolumn{15}{c}{UCR} \\ \cline{2-16}
&	\text{93.3}	&	\text{87.8}	&	\text{86.37}	&	\text{66.41}	&	\text{83.47}	&	\text{97.71}	&	\text{90.83}	&	\text{86.11}	&	\text{73.87}	&	\text{87.13}	&	\text{97.92}	&	\text{91.49}	&	\text{86.39}	&	\text{74.93}	&	\text{87.68}	\\

\midrule
\multirow{4}{*}{closed}
& \multicolumn{15}{c}{H-score/H$^3$-score} \\ \cline{2-16}
&	\text{73.51}	&	\text{69.37}	&	\text{70.11}	&	\text{52.58}	&	\text{66.39}	&	\text{82.0}	&	\text{76.95}	&	\text{68.83}	&	\text{58.82}	&	\text{71.65}	&	\text{83.34}	&	\text{77.76}	&	\text{74.03}	&	\text{60.16}	&	\text{73.82}	\\
\cline{2-16}
& \multicolumn{15}{c}{UCR} \\ \cline{2-16}
&	\text{79.25}	&	\text{83.89}	&	\text{87.31}	&	\text{74.02}	&	\text{81.12}	&	\text{87.35}	&	\text{89.17}	&	\text{87.56}	&	\text{79.58}	&	\text{85.91}	&	\text{87.88}	&	\text{89.71}	&	\text{88.11}	&	\text{80.33}	&	\text{86.51}	\\

\midrule
\multirow{4}{*}{partial}
& \multicolumn{15}{c}{H-score/H$^3$-score} \\ \cline{2-16}
&	\text{85.58}	&	\text{74.64}	&	\text{77.57}	&	\text{57.41}	&	\text{73.8}	&	\text{93.31}	&	\text{79.32}	&	\text{75.98}	&	\text{64.49}	&	\text{78.28}	&	\text{94.32}	&	\text{79.52}	&	\text{81.83}	&	\text{65.89}	&	\text{80.39}	\\
\cline{2-16}
& \multicolumn{15}{c}{UCR} \\ \cline{2-16}
&	\text{86.56}	&	\text{86.54}	&	\text{88.29}	&	\text{76.58}	&	\text{84.49}	&	\text{95.89}	&	\text{89.45}	&	\text{88.08}	&	\text{81.34}	&	\text{88.69}	&	\text{96.61}	&	\text{89.91}	&	\text{88.81}	&	\text{82.03}	&	\text{89.34}	\\
\bottomrule
\end{tabular}
\end{adjustbox}
\caption{Results of CLIP distillation based on different CLIP models.}
\label{table:results-different-backbones}
\end{table}

\section{Conclusion, limitations and future work}
In this paper, inspired by the robustness of large-scale pre-trained models to distribution shifts, we set out to develop a UniDA method utilizing these foundation models. We initially conducted comprehensive experiments to evaluate how the existing state-of-the-art UniDA methods perform when applied to foundation models. Our analysis of the results revealed several noteworthy findings, indicating the necessity for further research in the context of UniDA with foundation models.
As a response to these insights, we introduced a straightforward method involving target data distillation, which establishes a new state-of-the-art in UniDA using CLIP models. The significant improvements over previous results demonstrate the promising potential of employing foundation models for UniDA tasks. We hope that our investigation and the introduction of this straightforward framework can act as a robust baseline, thus promoting future research in this domain.

Our work has certain limitations. For instance, we focused on freezing the encoder when using foundation models due to the subpar results observed in full fine-tuning. However, recent studies have introduced new techniques for improving full fine-tuning with these models, such as the fine-tuning pre-trained methods \cite{goyal2022finetune} and surgical fine-tuning \cite{lee2023surgical}. We did not explore these techniques due to the substantial computational resources required, leaving this as a potential avenue for future research.
Furthermore, our method does not incorporate source data during the training process. We anticipate that future work can enhance our approach by leveraging information from the source data to further improve its performance.

\clearpage
\appendix
\renewcommand\thetable{\Alph{section}\arabic{table}}
\setcounter{table}{0}
\section{Experimental setup details}

\noindent \textbf{Dataset}:
We provide detail information about four datasets -- Office \cite{office2010adapting}, OfficeHome (OH) \cite{offhome2017deep}, VisDA (VD) \cite{visda2017visda}, and DomainNet (DN) \cite{domainnet2019moment} -- in Table \ref{tab:dataset-info}.

\begin{table}[ht]
    \centering
    \begin{adjustbox}{width=1.0\textwidth,center}
    \begin{tabular}{c|c|c|c|c|c}
    \toprule
        Office 	&	Domains	&	Amazon (A)	&	DSLR (D)	&	Webcam (W) & -	\\ \cline{2-6}
        (31 categories) 	                    &	Number of Samples	&	2817	&	498	&	795	 & - \\ \hline

        OfficeHome &   Domains &	Art (A)	&	Clipart (C)	&	Product (P)	&	RealWorld (R) \\ \cline{2-6}
        (65 categories)                            &   Number of Samples &	2427	&	4365	&	4439	&	4357 \\ \hline

        VisDA      &   Domains &  Syn (S)	& Real (R) & - & - \\ \cline{2-6}
        (12 categories)                            &   Number of Samples &  152397 &	55388 & - & - \\ \hline

        DomainNet	&	Domains  &	Painting (P)	&	Real (R)	&	Sketch (S)	& - \\ \cline{2-6}
        (345 categories)                            &   Number of Samples &	50416	&	120906	&	48212	& - \\

        \bottomrule
    \end{tabular}
    \end{adjustbox}
    \caption{Datasets information.}
    \label{tab:dataset-info}
\end{table}

\noindent \textbf{Classes split settings}:
The total categories of each dataset are split into the three disjoint parts -- common categories $\mathcal{Y}^{st}$, source private categories $\mathcal{Y}^{s/t}$, and target private categories $\mathcal{Y}^{t/s}$ -- to consist source and target domains. Since $|\mathcal{Y}^{st}|+|\mathcal{Y}^{s/t}|+|\mathcal{Y}^{t/s}|$ is fixed, we name each split setting as ($|\mathcal{Y}^{st}|/|\mathcal{Y}^{s/t}|$). The split settings for different datasets are shown in Table \ref{tab:split-settings}, following previous setting protocols \cite{saito2020universal}. Note that we only assign two split settings to DomainNet is because of the absence of samples in some categories in the Painting domain.
\begin{table}[ht]
    \centering
    \begin{tabular}{c|cccc} \toprule
         \multirow{2}{*}{Datasets} & \multicolumn{4}{c}{Split settings} \\ \cline{2-5}
                                & open-partial & open & closed & partial \\
         \midrule
         Office   & (10/10) & (10/0) & (31/0) & (10/21) \\
         OfficeHome & (10/5) & (15/0) & (65/0) & (25/40) \\
         VisDA    & (6/3) & (6/0) & (12/0) & (6/6) \\
         DomainNet & (150/50) & (150/0) & (345/0) & (150/195) \\
    \bottomrule
    \end{tabular}
    \caption{Classes split settings on four datasets.}
    \label{tab:split-settings}
\end{table}

\noindent \textbf{Number of training iterations}:
The maximum number of training iterations for the model is determined based on the scale of the training dataset. It is set to either 5000, 10000, or 20000 for different task settings, as indicated in Table \ref{tab:training-iterations}.
\begin{table}[ht]
    \centering
    \begin{tabular}{c|cccc} \toprule
         \multirow{2}{*}{Datasets} & \multicolumn{4}{c}{Split settings} \\ \cline{2-5}
                                & open-partial & open & closed & partial \\
         \midrule
         Office   & 5000 & 5000 & 10000 & 10000 \\
         OfficeHome & 5000 & 5000 & 10000 & 10000 \\
         VisDA    & 10000 & 10000 & 20000 & 20000 \\
         DomainNet & 10000 & 10000 & 20000 & 20000 \\
    \bottomrule
    \end{tabular}
    \caption{Training iterations on different task settings.}
    \label{tab:training-iterations}
\end{table}

\noindent \textbf{Text template using for CLIP zero-shot method}:
We follow the ensemble text templates in \cite{lin2023multimodality} for CLIP zero-shot method, which include 180 templates.
Each class prototype is calculated as the mean vector of the 180 corresponding text encoding vectors.

\noindent \textbf{Compute description}:
Our computing resource is a single GPU of NVIDIA GeForce RTX 3090 with 32 Intel(R) Xeon(R) Silver 4215R CPU @ 3.20GHz.

\noindent \textbf{Existing codes used}:
To fair comparison to different methods, we build a code farmework -- UniOOD, which integrates many previous methods. All codes to implement previous methods are directly copied from their official codes:

DANCE \cite{saito2020universal}: \url{https://github.com/VisionLearningGroup/DANCE};

OVANet \cite{saito2021ovanet}: \url{https://github.com/VisionLearningGroup/OVANet};

UniOT \cite{chang2022unified}: \url{https://github.com/changwxx/UniOT-for-UniDA};

WiSE-FT \cite{wortsman2022robust}: \url{https://github.com/mlfoundations/wise-ft};

CLIP cross-model \cite{lin2023multimodality}: \url{https://github.com/linzhiqiu/cross_modal_adaptation}.

The use of DINOv2 \cite{oquab2023dinov2} and CLIP models \cite{radford2021learning} follows \url{https://github.com/facebookresearch/dinov2} and \url{https://github.com/openai/CLIP} respectively.

\section{Detail experimental results}
\setcounter{table}{0}
\begin{table}[htp]
    \centering
    \begin{adjustbox}{width=1.0\textwidth,center}
\begin{tabular}{c|cccccc|c} \toprule
Methods	&	A2D	&	A2W	&	D2A	&	D2W	&	W2A	&	W2D	&	Avg	\\ \midrule
\multicolumn{8}{c}{H-score} \\ \midrule
SO	&	54.48$\pm$3.1	&	53.47$\pm$1.48	&	79.11$\pm$0.25	&	76.59$\pm$1.0	&	73.56$\pm$1.42	&	60.82$\pm$1.84	&	66.34	\\
DANCE	&	77.87$\pm$0.94	&	75.97$\pm$0.92	&	77.0$\pm$1.74	&	90.92$\pm$2.59	&	72.37$\pm$2.69	&	87.79$\pm$2.16	&	80.32	\\
OVANet	&	\textbf{82.95}$\pm$0.56	&	\textbf{77.61}$\pm$1.19	&	67.91$\pm$2.24	&	\textbf{94.99}$\pm$0.3	&	81.25$\pm$0.37	&	\textbf{95.26}$\pm$0.27	&	83.33	\\
UniOT	&	77.09$\pm$1.11	&	76.73$\pm$0.57	&	\textbf{86.44}$\pm$0.42	&	91.22$\pm$0.79	&	\textbf{85.45}$\pm$0.53	&	89.3$\pm$0.52	&	\textbf{84.37}	\\ \midrule

\multicolumn{8}{c}{H$^3$-score} \\ \midrule
SO	&	60.18$\pm$2.76	&	59.35$\pm$1.0	&	66.71$\pm$0.44	&	75.07$\pm$0.61	&	65.44$\pm$0.49	&	66.54$\pm$1.46	&	65.55	\\
DANCE	&	73.08$\pm$1.55	&	70.89$\pm$0.66	&	64.03$\pm$1.56	&	79.95$\pm$1.89	&	62.11$\pm$1.63	&	79.35$\pm$1.81	&	71.57	\\
OVANet	&	\textbf{80.45}$\pm$0.67	&	\textbf{76.76}$\pm$1.04	&	59.72$\pm$1.45	&	\textbf{87.29}$\pm$1.24	&	67.26$\pm$0.37	&	\textbf{89.29}$\pm$0.93	&	76.8	\\
UniOT	&	76.88$\pm$2.23	&	75.02$\pm$0.64	&	\textbf{69.41}$\pm$1.86	&	86.17$\pm$1.65	&	\textbf{68.86}$\pm$1.4	&	87.66$\pm$0.74	&	\textbf{77.33}	\\ \midrule

\multicolumn{8}{c}{UCR} \\ \midrule
SO	&	69.72$\pm$1.45	&	62.83$\pm$2.31	&	79.54$\pm$0.62	&	94.19$\pm$1.27	&	82.78$\pm$0.9	&	98.2$\pm$0.22	&	81.21	\\
DANCE	&	\textbf{79.04}$\pm$2.5	&	\textbf{79.86}$\pm$0.87	&	82.61$\pm$0.48	&	93.21$\pm$1.29	&	81.68$\pm$0.72	&	90.42$\pm$2.18	&	84.47	\\
OVANet	&	71.79$\pm$0.64	&	65.18$\pm$1.03	&	73.98$\pm$2.03	&	\textbf{97.3}$\pm$0.67	&	81.46$\pm$0.44	&	\textbf{98.54}$\pm$0.14	&	81.38	\\
UniOT	&	72.66$\pm$2.24	&	72.81$\pm$2.18	&	\textbf{87.12}$\pm$0.98	&	94.74$\pm$0.75	&	\textbf{87.51}$\pm$0.5	&	93.57$\pm$2.33	&	\textbf{84.73}	\\
\bottomrule
    \end{tabular}
    \end{adjustbox}
    \caption{Office: ResNet50 \& (10/10) setting}
    \label{tab:office-resnet50-10-10}
\end{table}

\begin{table}[htp]
    \centering
    \begin{adjustbox}{width=1.0\textwidth,center}
    \begin{tabular}{c|cccccc|c} \toprule
Methods	&	A2D	&	A2W	&	D2A	&	D2W	&	W2A	&	W2D	&	Avg	\\ \midrule
\multicolumn{8}{c}{H-score} \\ \midrule
SO	&	89.62$\pm$0.15	&	82.87$\pm$0.1	&	\textbf{89.1}$\pm$0.36	&	94.71$\pm$0.4	&	88.53$\pm$0.59	&	92.74$\pm$0.56	&	89.6	\\
DANCE	&	\textbf{91.38}$\pm$0.26	&	\textbf{86.19}$\pm$0.91	&	87.33$\pm$1.86	&	\textbf{99.56}$\pm$0.18	&	\textbf{89.14}$\pm$2.89	&	\textbf{97.96}$\pm$0.42	&	\textbf{91.93}	\\
OVANet	&	82.41$\pm$5.27	&	81.95$\pm$1.12	&	88.62$\pm$0.31	&	90.59$\pm$4.13	&	87.74$\pm$2.5	&	87.75$\pm$2.88	&	86.51	\\
UniOT	&	82.89$\pm$0.24	&	84.82$\pm$2.02	&	88.29$\pm$0.3	&	94.79$\pm$1.23	&	87.1$\pm$0.94	&	97.09$\pm$0.61	&	89.16	\\ \midrule

\multicolumn{8}{c}{H$^3$-score} \\ \midrule
SO	&	89.26$\pm$0.1	&	86.9$\pm$0.07	&	\textbf{85.05}$\pm$0.22	&	95.23$\pm$0.27	&	84.7$\pm$0.36	&	91.3$\pm$0.36	&	88.74	\\
DANCE	&	\textbf{90.41}$\pm$0.17	&	\textbf{89.31}$\pm$0.65	&	83.96$\pm$1.14	&	\textbf{98.45}$\pm$0.12	&	\textbf{85.05}$\pm$1.77	&	94.6$\pm$0.26	&	\textbf{90.3}	\\
OVANet	&	84.28$\pm$3.73	&	86.22$\pm$0.82	&	84.76$\pm$0.19	&	92.37$\pm$2.9	&	84.2$\pm$1.55	&	87.99$\pm$1.94	&	86.64	\\
UniOT	&	85.77$\pm$1.1	&	87.65$\pm$2.43	&	81.74$\pm$1.02	&	94.21$\pm$0.58	&	80.72$\pm$0.27	&	\textbf{95.98}$\pm$0.45	&	87.68	\\ \midrule

\multicolumn{8}{c}{UCR} \\ \midrule
SO	&	\textbf{93.29}$\pm$0.26	&	\textbf{85.18}$\pm$0.16	&	83.24$\pm$0.33	&	99.07$\pm$0.51	&	\textbf{86.48}$\pm$0.62	&	\textbf{99.91}$\pm$0.0	&	\textbf{91.2}	\\
DANCE	&	84.39$\pm$0.16	&	79.17$\pm$0.45	&	78.92$\pm$5.46	&	\textbf{100.0}$\pm$0.0	&	81.83$\pm$5.82	&	99.61$\pm$0.24	&	87.32	\\
OVANet	&	92.04$\pm$2.44	&	83.05$\pm$0.68	&	83.01$\pm$1.05	&	96.28$\pm$2.62	&	86.31$\pm$1.18	&	99.05$\pm$0.27	&	89.96	\\
UniOT	&	77.59$\pm$2.95	&	82.07$\pm$2.75	&	\textbf{83.98}$\pm$1.7	&	94.6$\pm$1.0	&	82.89$\pm$2.59	&	98.67$\pm$0.19	&	86.63	\\
\bottomrule
    \end{tabular}
    \end{adjustbox}
    \caption{Office: DINOv2 \& (10/10) setting}
    \label{tab:office-dinov2-10-10}
\end{table}
\begin{table}[htp]
    \centering
    \begin{adjustbox}{width=1.0\textwidth,center}
    \begin{tabular}{c|cccccc|c} \toprule
Methods	&	A2D	&	A2W	&	D2A	&	D2W	&	W2A	&	W2D	&	Avg	\\ \midrule
\multicolumn{8}{c}{H-score} \\ \midrule
SO	&	92.71$\pm$0.13	&	89.3$\pm$0.19	&	90.08$\pm$0.21	&	93.95$\pm$0.17	&	88.16$\pm$0.22	&	97.69$\pm$0.2	&	91.98	\\
DANCE	&	\textbf{96.02}$\pm$0.2	&	90.18$\pm$0.83	&	93.68$\pm$2.56	&	\textbf{98.64}$\pm$0.21	&	90.23$\pm$2.67	&	\textbf{99.42}$\pm$0.24	&	\textbf{94.69}	\\
OVANet	&	93.82$\pm$0.58	&	88.88$\pm$0.63	&	92.3$\pm$2.09	&	97.63$\pm$0.0	&	89.35$\pm$0.21	&	98.16$\pm$0.0	&	93.36	\\
UniOT	&	84.59$\pm$1.64	&	\textbf{92.24}$\pm$1.23	&	\textbf{94.5}$\pm$1.58	&	94.84$\pm$1.79	&	\textbf{94.76}$\pm$0.95	&	92.99$\pm$0.61	&	92.32	\\ \hline
WiSE-FT	&	77.89$\pm$0.42	&	70.19$\pm$0.1	&	80.83$\pm$0.22	&	92.68$\pm$0.17	&	76.75$\pm$0.33	&	95.7$\pm$0.37	&	82.34	\\
CLIP cross-model	&	94.24$\pm$0.39	&	89.82$\pm$0.32	&	92.34$\pm$0.11	&	93.65$\pm$0.3	&	92.23$\pm$0.05	&	95.99$\pm$0.24	&	93.05	\\
\bottomrule
    \end{tabular}
    \end{adjustbox}
    \caption{Office: CLIP \& (10/10) setting}
    \label{tab:office-clip-10-10}
\end{table}

\begin{table}[htp]
    \centering
    \begin{adjustbox}{width=1.0\textwidth,center}
    \begin{tabular}{c|cccccccccccc|c} \toprule
Methods	&	A2C	&	A2P	&	A2R	&	C2A	&	C2P	&	C2R	&	P2A	&	P2C	&	P2R	&	R2A	&	R2C	&	R2P	&	Avg	\\ \midrule
\multicolumn{14}{c}{H-score} \\ \midrule
SO	&	50.35$\pm$0.25	&	50.87$\pm$0.31	&	55.44$\pm$0.59	&	59.31$\pm$1.19	&	49.15$\pm$0.57	&	57.57$\pm$1.09	&	62.01$\pm$0.48	&	50.21$\pm$0.68	&	56.83$\pm$0.41	&	56.43$\pm$0.12	&	52.31$\pm$0.3	&	52.19$\pm$0.41	&	54.39	\\
DANCE	&	39.64$\pm$3.7	&	38.23$\pm$5.91	&	38.69$\pm$3.66	&	38.55$\pm$3.4	&	13.22$\pm$0.27	&	37.21$\pm$0.43	&	51.73$\pm$2.72	&	43.89$\pm$1.31	&	43.2$\pm$1.78	&	29.33$\pm$4.01	&	44.47$\pm$3.19	&	50.62$\pm$0.88	&	39.06	\\
OVANet	&	58.01$\pm$0.64	&	78.91$\pm$0.18	&	82.15$\pm$0.56	&	69.4$\pm$0.63	&	68.1$\pm$0.22	&	76.41$\pm$0.08	&	71.98$\pm$0.58	&	56.77$\pm$0.23	&	81.72$\pm$0.11	&	\textbf{77.94}$\pm$0.55	&	58.91$\pm$0.21	&	79.81$\pm$0.46	&	71.68	\\
UniOT	&	\textbf{66.13}$\pm$0.97	&	\textbf{80.42}$\pm$0.7	&	\textbf{84.56}$\pm$0.54	&	\textbf{72.79}$\pm$0.2	&	\textbf{76.59}$\pm$1.66	&	\textbf{82.42}$\pm$0.96	&	\textbf{75.82}$\pm$0.96	&	\textbf{65.87}$\pm$0.76	&	\textbf{85.07}$\pm$1.14	&	76.61$\pm$0.3	&	\textbf{64.8}$\pm$1.23	&	\textbf{80.6}$\pm$0.44	&	\textbf{75.97}	\\ \midrule

\multicolumn{14}{c}{H$^3$-score} \\ \midrule
SO	&	51.29$\pm$0.3	&	57.77$\pm$0.29	&	60.11$\pm$0.52	&	57.45$\pm$0.76	&	55.1$\pm$0.45	&	60.21$\pm$0.78	&	60.58$\pm$0.38	&	50.49$\pm$0.28	&	61.05$\pm$0.32	&	57.72$\pm$0.08	&	52.1$\pm$0.12	&	58.55$\pm$0.33	&	56.87	\\
\bottomrule
    \end{tabular}
    \end{adjustbox}
    \caption{OfficeHome: ResNet50 \& (10/5) setting}
    \label{tab:officehome-resnet50-10-5}
\end{table}

\begin{table}[htp]
    \centering
    \begin{adjustbox}{width=1.0\textwidth,center}
    \begin{tabular}{c|cccccccccccc|c} \toprule
Methods	&	A2C	&	A2P	&	A2R	&	C2A	&	C2P	&	C2R	&	P2A	&	P2C	&	P2R	&	R2A	&	R2C	&	R2P	&	Avg	\\ \midrule
\multicolumn{14}{c}{H-score} \\ \midrule
SO	&	73.23$\pm$0.2	&	86.77$\pm$0.25	&	88.11$\pm$0.18	&	85.08$\pm$0.15	&	86.37$\pm$0.09	&	89.68$\pm$0.16	&	81.99$\pm$0.24	&	71.77$\pm$0.25	&	85.24$\pm$0.22	&	83.15$\pm$0.27	&	75.66$\pm$0.23	&	86.15$\pm$0.15	&	82.77	\\
DANCE	&	79.49$\pm$0.68	&	84.7$\pm$0.29	&	89.82$\pm$0.58	&	85.41$\pm$0.22	&	90.3$\pm$0.58	&	91.36$\pm$0.12	&	80.89$\pm$0.64	&	70.7$\pm$1.14	&	85.46$\pm$0.47	&	85.01$\pm$0.24	&	77.72$\pm$0.29	&	91.74$\pm$0.33	&	84.38	\\
OVANet	&	72.18$\pm$0.25	&	83.0$\pm$0.17	&	82.62$\pm$0.45	&	75.83$\pm$0.57	&	77.15$\pm$0.77	&	72.46$\pm$1.4	&	82.0$\pm$0.15	&	56.85$\pm$0.33	&	85.45$\pm$0.17	&	82.2$\pm$0.8	&	66.38$\pm$0.12	&	85.79$\pm$0.22	&	76.83	\\
UniOT	&	\textbf{82.07}$\pm$0.74	&	\textbf{92.04}$\pm$1.57	&	\textbf{93.06}$\pm$0.43	&	\textbf{86.88}$\pm$0.37	&	\textbf{90.87}$\pm$2.11	&	\textbf{91.93}$\pm$0.18	&	\textbf{83.8}$\pm$0.52	&	\textbf{77.44}$\pm$1.85	&	\textbf{90.22}$\pm$0.49	&	\textbf{85.65}$\pm$0.61	&	\textbf{81.24}$\pm$0.59	&	\textbf{95.29}$\pm$0.26	&	\textbf{87.54}	\\ \midrule

\multicolumn{14}{c}{H$^3$-score} \\ \midrule
SO	&	74.52$\pm$0.14	&	87.67$\pm$0.17	&	87.28$\pm$0.11	&	83.16$\pm$0.09	&	87.4$\pm$0.06	&	88.31$\pm$0.1	&	81.17$\pm$0.16	&	73.5$\pm$0.18	&	85.38$\pm$0.15	&	81.93$\pm$0.17	&	76.18$\pm$0.15	&	87.25$\pm$0.1	&	82.81	\\
\bottomrule
    \end{tabular}
    \end{adjustbox}
    \caption{OfficeHome: DINOv2 \& (10/5) setting}
    \label{tab:officehome-dinov2-10-5}
\end{table}

\begin{table}[htp]
    \centering
    \begin{adjustbox}{width=1.0\textwidth,center}
    \begin{tabular}{c|cccccccccccc|c} \toprule
Methods	&	A2C	&	A2P	&	A2R	&	C2A	&	C2P	&	C2R	&	P2A	&	P2C	&	P2R	&	R2A	&	R2C	&	R2P	&	Avg	\\ \midrule
\multicolumn{14}{c}{H-score} \\ \midrule
SO	&	75.73$\pm$0.08	&	84.45$\pm$0.07	&	88.31$\pm$0.02	&	86.28$\pm$0.16	&	88.14$\pm$0.23	&	91.1$\pm$0.12	&	81.52$\pm$0.14	&	78.42$\pm$0.1	&	89.53$\pm$0.13	&	85.3$\pm$0.07	&	81.68$\pm$0.04	&	83.82$\pm$0.09	&	84.52	\\
DANCE	&	82.2$\pm$0.08	&	\textbf{94.02}$\pm$0.01	&	90.2$\pm$0.06	&	86.63$\pm$0.16	&	\textbf{93.72}$\pm$0.16	&	\textbf{93.79}$\pm$0.02	&	85.35$\pm$0.21	&	84.47$\pm$0.09	&	91.7$\pm$0.11	&	86.44$\pm$0.11	&	83.97$\pm$0.05	&	\textbf{95.64}$\pm$0.39	&	89.01	\\
OVANet	&	80.55$\pm$0.4	&	91.84$\pm$0.08	&	90.74$\pm$0.27	&	86.07$\pm$0.17	&	90.52$\pm$0.11	&	91.78$\pm$0.12	&	75.64$\pm$0.67	&	68.54$\pm$0.71	&	90.29$\pm$0.12	&	85.74$\pm$0.2	&	80.56$\pm$0.31	&	92.75$\pm$0.77	&	85.42	\\
UniOT	&	\textbf{88.28}$\pm$0.33	&	92.03$\pm$0.56	&	\textbf{93.39}$\pm$0.31	&	87.61$\pm$0.56	&	90.19$\pm$0.44	&	92.09$\pm$0.82	&	85.99$\pm$0.37	&	\textbf{86.08}$\pm$0.36	&	92.04$\pm$0.7	&	86.87$\pm$0.13	&	\textbf{86.47}$\pm$0.43	&	92.35$\pm$0.57	&	\textbf{89.45}	\\ \hline
WiSE-FT	&	68.39$\pm$0.04	&	89.68$\pm$0.01	&	90.9$\pm$0.04	&	69.46$\pm$0.27	&	87.43$\pm$0.14	&	87.14$\pm$0.2	&	68.7$\pm$0.45	&	59.92$\pm$0.34	&	85.8$\pm$0.05	&	78.64$\pm$0.26	&	73.97$\pm$0.29	&	92.43$\pm$0.11	&	79.37	\\
CLIP cross-model	&	78.37$\pm$0.06	&	85.06$\pm$0.05	&	89.18$\pm$0.12	&	88.98$\pm$0.08	&	88.09$\pm$0.24	&	91.52$\pm$0.1	&	85.91$\pm$0.05	&	82.6$\pm$0.11	&	91.59$\pm$0.07	&	86.8$\pm$0.17	&	82.09$\pm$0.05	&	84.24$\pm$0.09	&	86.2	\\
\bottomrule
    \end{tabular}
    \end{adjustbox}
    \caption{OfficeHome: CLIP \& (10/5) setting}
    \label{tab:officehome-clip-10-5}
\end{table}

\begin{table}[htp]
    \centering
    \begin{adjustbox}{width=1.0\textwidth,center}
    \begin{tabular}{c|cccccc|c} \toprule
Methods	&	P2R	&	P2S	&	R2P	&	R2S	&	S2P	&	S2R	&	Avg	\\ \midrule
\multicolumn{8}{c}{H-score} \\ \midrule
SO	&	42.24$\pm$0.13	&	40.84$\pm$0.22	&	45.31$\pm$0.43	&	41.1$\pm$0.12	&	28.05$\pm$0.24	&	37.67$\pm$0.47	&	39.2	\\
DANCE	&	20.09$\pm$0.26	&	25.86$\pm$0.92	&	34.6$\pm$1.19	&	41.83$\pm$0.42	&	18.68$\pm$0.49	&	20.42$\pm$0.29	&	26.91	\\
OVANet	&	55.29$\pm$0.08	&	45.35$\pm$0.2	&	\textbf{52.17}$\pm$0.17	&	44.69$\pm$0.05	&	44.62$\pm$0.33	&	55.32$\pm$0.06	&	49.57	\\
UniOT	&	\textbf{56.75}$\pm$0.11	&	\textbf{47.4}$\pm$0.21	&	51.72$\pm$0.23	&	\textbf{47.23}$\pm$0.24	&	\textbf{46.17}$\pm$0.23	&	\textbf{56.02}$\pm$0.28	&	\textbf{50.88}	\\ \midrule

\multicolumn{8}{c}{H$^3$-score} \\ \midrule
SO	&	47.97$\pm$0.11	&	40.92$\pm$0.13	&	47.43$\pm$0.29	&	41.54$\pm$0.08	&	31.99$\pm$0.21	&	43.08$\pm$0.39	&	42.16	\\
DANCE	&	26.22$\pm$0.31	&	29.49$\pm$0.86	&	38.98$\pm$1.04	&	42.53$\pm$0.33	&	23.42$\pm$0.54	&	26.47$\pm$0.31	&	31.18	\\
OVANet	&	\textbf{58.24}$\pm$0.07	&	\textbf{44.2}$\pm$0.15	&	\textbf{52.12}$\pm$0.12	&	\textbf{43.67}$\pm$0.08	&	\textbf{46.09}$\pm$0.23	&	\textbf{57.72}$\pm$0.1	&	\textbf{50.34}	\\
UniOT	&	55.7$\pm$0.12	&	42.39$\pm$0.18	&	47.78$\pm$0.28	&	42.51$\pm$0.18	&	44.28$\pm$0.17	&	54.78$\pm$0.29	&	47.91	\\ \midrule

\multicolumn{8}{c}{UCR} \\ \midrule
SO	&	43.03$\pm$0.23	&	25.57$\pm$0.21	&	36.19$\pm$0.13	&	26.03$\pm$0.15	&	21.6$\pm$0.02	&	33.85$\pm$0.31	&	31.04	\\
DANCE	&	42.56$\pm$0.13	&	23.88$\pm$0.58	&	35.89$\pm$0.58	&	\textbf{29.33}$\pm$0.36	&	25.61$\pm$0.12	&	38.99$\pm$0.09	&	32.71	\\
OVANet	&	43.03$\pm$0.11	&	27.57$\pm$0.38	&	\textbf{38.18}$\pm$0.02	&	27.65$\pm$0.26	&	\textbf{30.79}$\pm$0.29	&	39.34$\pm$0.19	&	34.43	\\
UniOT	&	\textbf{43.41}$\pm$0.15	&	\textbf{27.72}$\pm$0.42	&	36.16$\pm$0.06	&	28.57$\pm$0.34	&	29.8$\pm$0.09	&	\textbf{41.48}$\pm$0.43	&	\textbf{34.52}	\\
\bottomrule
    \end{tabular}
    \end{adjustbox}
    \caption{DomainNet: ResNet50 \& (150/50) setting}
    \label{tab:domainnet-resnet50-150-50}
\end{table}

\begin{table}[htp]
    \centering
    \begin{adjustbox}{width=1.0\textwidth,center}
    \begin{tabular}{c|cccccc|c} \toprule
Methods	&	P2R	&	P2S	&	R2P	&	R2S	&	S2P	&	S2R	&	Avg	\\ \midrule
\multicolumn{8}{c}{H-score} \\ \midrule
SO	&	71.74$\pm$0.06	&	68.13$\pm$0.08	&	67.37$\pm$0.1	&	67.5$\pm$0.17	&	64.85$\pm$0.13	&	70.71$\pm$0.07	&	68.38	\\
DANCE	&	71.64$\pm$0.12	&	68.86$\pm$0.17	&	67.64$\pm$0.05	&	68.76$\pm$0.15	&	65.19$\pm$0.08	&	70.36$\pm$0.18	&	68.74	\\
OVANet	&	51.66$\pm$0.49	&	66.05$\pm$0.02	&	66.69$\pm$0.19	&	64.24$\pm$0.1	&	53.8$\pm$0.41	&	32.11$\pm$0.45	&	55.76	\\
UniOT	&	\textbf{73.44}$\pm$0.28	&	\textbf{68.89}$\pm$0.16	&	\textbf{67.94}$\pm$0.07	&	\textbf{70.16}$\pm$0.18	&	\textbf{65.6}$\pm$0.43	&	\textbf{73.1}$\pm$0.04	&	\textbf{69.86}	\\ \midrule

\multicolumn{8}{c}{H$^3$-score} \\ \midrule
SO	&	\textbf{76.17}$\pm$0.04	&	67.99$\pm$0.05	&	70.07$\pm$0.08	&	67.57$\pm$0.11	&	68.22$\pm$0.1	&	75.39$\pm$0.05	&	70.9	\\
DANCE	&	76.09$\pm$0.09	&	\textbf{68.47}$\pm$0.11	&	\textbf{70.26}$\pm$0.04	&	\textbf{68.4}$\pm$0.1	&	\textbf{68.48}$\pm$0.06	&	75.13$\pm$0.14	&	\textbf{71.14}	\\
OVANet	&	59.74$\pm$0.44	&	66.59$\pm$0.02	&	69.58$\pm$0.14	&	65.36$\pm$0.07	&	59.63$\pm$0.34	&	40.65$\pm$0.49	&	60.26	\\
UniOT	&	75.69$\pm$0.18	&	66.74$\pm$0.08	&	69.12$\pm$0.04	&	68.12$\pm$0.2	&	67.24$\pm$0.26	&	\textbf{75.67}$\pm$0.11	&	70.43	\\ \midrule

\multicolumn{8}{c}{UCR} \\ \midrule
SO	&	67.3$\pm$0.13	&	57.35$\pm$0.15	&	60.28$\pm$0.09	&	57.72$\pm$0.26	&	58.33$\pm$0.09	&	67.72$\pm$0.04	&	61.45	\\
DANCE	&	\textbf{67.98}$\pm$0.22	&	\textbf{60.11}$\pm$0.14	&	\textbf{62.85}$\pm$0.07	&	\textbf{61.64}$\pm$0.08	&	\textbf{60.58}$\pm$0.2	&	\textbf{67.96}$\pm$0.22	&	\textbf{63.52}	\\
OVANet	&	64.04$\pm$0.05	&	53.42$\pm$0.09	&	58.34$\pm$0.14	&	53.91$\pm$0.12	&	56.9$\pm$0.2	&	64.97$\pm$0.08	&	58.6	\\
UniOT	&	64.18$\pm$0.33	&	53.61$\pm$0.19	&	54.84$\pm$0.16	&	56.26$\pm$0.42	&	52.74$\pm$0.45	&	65.45$\pm$0.18	&	57.85	\\
\bottomrule
    \end{tabular}
    \end{adjustbox}
    \caption{DomainNet: DINOv2 \& (150/50) setting}
    \label{tab:domainnet-dinov2-150-50}
\end{table}

\begin{table}[htp]
    \centering
    \begin{adjustbox}{width=1.0\textwidth,center}
    \begin{tabular}{c|cccccc|c} \toprule
Methods	&	P2R	&	P2S	&	R2P	&	R2S	&	S2P	&	S2R	&	Avg	\\ \midrule
\multicolumn{8}{c}{H-score} \\ \midrule
SO	&	67.35$\pm$0.07	&	57.55$\pm$0.04	&	58.17$\pm$0.12	&	61.14$\pm$0.19	&	53.01$\pm$0.14	&	71.72$\pm$0.06	&	61.49	\\
DANCE	&	67.6$\pm$0.07	&	57.33$\pm$0.06	&	55.57$\pm$0.25	&	59.65$\pm$0.06	&	52.09$\pm$0.08	&	70.94$\pm$0.12	&	60.53	\\
OVANet	&	74.87$\pm$0.15	&	69.55$\pm$0.11	&	67.97$\pm$0.27	&	70.41$\pm$0.17	&	65.73$\pm$0.09	&	75.65$\pm$0.11	&	70.7	\\
UniOT	&	74.67$\pm$0.39	&	\textbf{69.56}$\pm$0.19	&	\textbf{69.32}$\pm$0.47	&	71.38$\pm$0.21	&	67.15$\pm$0.52	&	76.42$\pm$0.16	&	71.42	\\ \hline
WiSE-FT	&	5.79$\pm$0.04	&	1.51$\pm$0.04	&	4.24$\pm$0.14	&	3.71$\pm$0.03	&	1.77$\pm$0.04	&	5.45$\pm$0.09	&	3.74	\\
CLIP cross-model	&	69.19$\pm$0.1	&	58.18$\pm$0.14	&	57.11$\pm$0.03	&	60.85$\pm$0.18	&	53.13$\pm$0.16	&	73.43$\pm$0.05	&	61.98	\\
\bottomrule
    \end{tabular}
    \end{adjustbox}
    \caption{DomainNet: CLIP \& (150/50) setting}
    \label{tab:domainnet-clip-150-50}
\end{table}

\begin{table}[htp]
    \centering
    \begin{adjustbox}{width=1.0\textwidth,center}
    \begin{tabular}{c|cccccc|c} \toprule
Methods	&	A2D	&	A2W	&	D2A	&	D2W	&	W2A	&	W2D	&	Avg	\\ \midrule
\multicolumn{8}{c}{H-score} \\ \midrule
SO	&	95.4$\pm$0.19	&	93.49$\pm$0.03	&	88.88$\pm$0.21	&	93.46$\pm$0.71	&	90.11$\pm$0.29	&	92.37$\pm$0.82	&	92.29	\\
DANCE	&	\textbf{96.5}$\pm$0.14	&	\textbf{96.13}$\pm$0.21	&	89.95$\pm$0.61	&	97.92$\pm$0.64	&	90.76$\pm$0.46	&	98.61$\pm$0.8	&	\textbf{94.98}	\\
OVANet	&	88.37$\pm$5.41	&	92.18$\pm$0.39	&	83.36$\pm$2.25	&	94.15$\pm$1.29	&	85.06$\pm$2.35	&	89.07$\pm$4.13	&	88.7	\\
UniOT	&	92.97$\pm$2.72	&	94.82$\pm$1.3	&	\textbf{90.52}$\pm$0.62	&	\textbf{97.97}$\pm$0.41	&	\textbf{90.96}$\pm$0.59	&	\textbf{99.9}$\pm$0.14	&	94.52	\\ \midrule

\multicolumn{8}{c}{H$^3$-score} \\ \midrule
SO	&	94.9$\pm$0.12	&	93.11$\pm$0.02	&	84.83$\pm$0.13	&	93.09$\pm$0.47	&	85.57$\pm$0.17	&	92.87$\pm$0.55	&	90.73	\\
DANCE	&	\textbf{95.62}$\pm$0.09	&	\textbf{94.84}$\pm$0.14	&	\textbf{85.47}$\pm$0.37	&	96.0$\pm$0.41	&	\textbf{85.96}$\pm$0.28	&	96.99$\pm$0.52	&	\textbf{92.48}	\\
OVANet	&	90.07$\pm$3.8	&	92.24$\pm$0.26	&	81.38$\pm$1.42	&	93.55$\pm$0.85	&	82.46$\pm$1.47	&	90.59$\pm$2.86	&	88.38	\\
UniOT	&	92.52$\pm$1.89	&	94.19$\pm$0.7	&	84.88$\pm$0.73	&	\textbf{96.83}$\pm$0.14	&	85.51$\pm$0.96	&	\textbf{97.49}$\pm$0.29	&	91.9	\\ \midrule

\multicolumn{8}{c}{UCR} \\ \midrule
SO	&	95.54$\pm$0.42	&	94.09$\pm$0.34	&	84.6$\pm$0.12	&	99.48$\pm$0.03	&	89.88$\pm$0.22	&	99.96$\pm$0.0	&	93.93	\\
DANCE	&	\textbf{97.23}$\pm$0.02	&	\textbf{95.83}$\pm$0.48	&	86.24$\pm$1.26	&	\textbf{99.71}$\pm$0.34	&	86.52$\pm$1.07	&	99.94$\pm$0.09	&	94.25	\\
OVANet	&	93.18$\pm$1.39	&	93.55$\pm$0.41	&	83.22$\pm$0.74	&	97.97$\pm$1.0	&	88.57$\pm$1.37	&	98.66$\pm$0.34	&	92.53	\\
UniOT	&	96.74$\pm$2.16	&	94.86$\pm$0.71	&	\textbf{90.98}$\pm$2.56	&	99.08$\pm$0.37	&	\textbf{91.39}$\pm$1.31	&	\textbf{100.0}$\pm$0.0	&	\textbf{95.51}	\\
\bottomrule
    \end{tabular}
    \end{adjustbox}
    \caption{Office: DINOv2 \& (10/0) setting}
    \label{tab:office-dinov2-10-0}
\end{table}

\begin{table}[htp]
    \centering
    \begin{adjustbox}{width=1.0\textwidth,center}
    \begin{tabular}{c|cccccc|c} \toprule
Methods	&	A2D	&	A2W	&	D2A	&	D2W	&	W2A	&	W2D	&	Avg	\\ \midrule
\multicolumn{8}{c}{H-score} \\ \midrule
SO	&	92.04$\pm$0.19	&	92.01$\pm$0.13	&	91.71$\pm$0.27	&	95.51$\pm$0.09	&	89.77$\pm$0.33	&	90.17$\pm$0.43	&	91.87	\\
DANCE	&	93.15$\pm$0.21	&	93.08$\pm$0.05	&	\textbf{96.74}$\pm$0.06	&	\textbf{98.74}$\pm$0.05	&	\textbf{95.81}$\pm$1.5	&	\textbf{99.0}$\pm$0.89	&	96.09	\\
OVANet	&	93.3$\pm$1.17	&	90.3$\pm$0.84	&	82.93$\pm$3.14	&	97.48$\pm$0.11	&	85.37$\pm$0.83	&	97.58$\pm$1.61	&	91.16	\\
UniOT	&	\textbf{95.11}$\pm$1.4	&	\textbf{95.92}$\pm$0.42	&	95.45$\pm$0.71	&	98.36$\pm$1.04	&	95.68$\pm$0.94	&	98.35$\pm$0.17	&	\textbf{96.48}	\\ \hline
WiSE-FT	&	92.77$\pm$0.0	&	90.99$\pm$0.0	&	92.78$\pm$0.37	&	97.87$\pm$0.08	&	92.1$\pm$0.23	&	97.93$\pm$0.2	&	94.07	\\
CLIP cross-model	&	93.51$\pm$0.16	&	92.19$\pm$0.05	&	93.85$\pm$0.01	&	94.97$\pm$0.0	&	92.47$\pm$0.01	&	94.54$\pm$0.08	&	93.59	\\
\bottomrule
    \end{tabular}
    \end{adjustbox}
    \caption{Office: CLIP \& (10/0) setting}
    \label{tab:office-clip-10-0}
\end{table}

\begin{table}[htp]
    \centering
    \begin{adjustbox}{width=1.0\textwidth,center}
    \begin{tabular}{c|cccccccccccc|c} \toprule
Methods	&	A2C	&	A2P	&	A2R	&	C2A	&	C2P	&	C2R	&	P2A	&	P2C	&	P2R	&	R2A	&	R2C	&	R2P	&	Avg	\\ \midrule
\multicolumn{14}{c}{H-score} \\ \midrule
SO	&	74.9$\pm$0.17	&	85.61$\pm$0.19	&	87.6$\pm$0.12	&	82.45$\pm$0.19	&	86.13$\pm$0.09	&	86.4$\pm$0.12	&	76.33$\pm$0.39	&	71.41$\pm$0.16	&	83.55$\pm$0.18	&	83.04$\pm$0.39	&	77.68$\pm$0.14	&	85.98$\pm$0.16	&	81.76	\\
DANCE	&	78.61$\pm$0.67	&	81.8$\pm$0.17	&	89.25$\pm$0.32	&	82.14$\pm$0.23	&	88.13$\pm$0.53	&	87.27$\pm$0.06	&	69.35$\pm$1.04	&	70.36$\pm$0.8	&	83.09$\pm$0.15	&	84.47$\pm$0.09	&	77.83$\pm$0.15	&	90.9$\pm$0.26	&	81.93	\\
OVANet	&	73.36$\pm$0.11	&	82.65$\pm$0.11	&	81.65$\pm$0.47	&	74.49$\pm$0.3	&	77.98$\pm$0.67	&	71.38$\pm$1.27	&	77.13$\pm$0.35	&	57.0$\pm$0.44	&	84.02$\pm$0.37	&	80.74$\pm$0.49	&	68.85$\pm$0.04	&	85.1$\pm$0.19	&	76.2	\\
UniOT	&	\textbf{80.59}$\pm$0.42	&	\textbf{87.84}$\pm$1.19	&	\textbf{91.77}$\pm$0.37	&	\textbf{84.98}$\pm$0.41	&	\textbf{89.74}$\pm$1.81	&	\textbf{88.21}$\pm$0.21	&	\textbf{80.2}$\pm$0.75	&	\textbf{76.9}$\pm$1.57	&	\textbf{86.74}$\pm$0.71	&	\textbf{86.03}$\pm$0.52	&	\textbf{80.99}$\pm$0.49	&	\textbf{92.69}$\pm$0.28	&	\textbf{85.56}	\\ \midrule

\multicolumn{14}{c}{H$^3$-score} \\ \midrule
SO	&	75.67$\pm$0.12	&	86.88$\pm$0.13	&	86.95$\pm$0.08	&	81.47$\pm$0.12	&	87.24$\pm$0.06	&	86.16$\pm$0.08	&	77.38$\pm$0.27	&	73.25$\pm$0.11	&	84.25$\pm$0.12	&	81.86$\pm$0.25	&	77.53$\pm$0.09	&	87.13$\pm$0.11	&	82.15	\\
\bottomrule
    \end{tabular}
    \end{adjustbox}
    \caption{OfficeHome: DINOv2 \& (15/0) setting}
    \label{tab:officehome-dinov2-15-0}
\end{table}

\begin{table}[htp]
    \centering
    \begin{adjustbox}{width=1.0\textwidth,center}
    \begin{tabular}{c|cccccccccccc|c} \toprule
Methods	&	A2C	&	A2P	&	A2R	&	C2A	&	C2P	&	C2R	&	P2A	&	P2C	&	P2R	&	R2A	&	R2C	&	R2P	&	Avg	\\ \midrule
\multicolumn{14}{c}{H-score} \\ \midrule
SO	&	75.41$\pm$0.09	&	83.2$\pm$0.05	&	86.19$\pm$0.14	&	81.69$\pm$0.12	&	87.15$\pm$0.21	&	86.87$\pm$0.08	&	75.93$\pm$0.28	&	76.58$\pm$0.06	&	86.49$\pm$0.21	&	83.02$\pm$0.09	&	78.68$\pm$0.02	&	83.39$\pm$0.08	&	82.05	\\
DANCE	&	80.03$\pm$0.04	&	89.48$\pm$0.04	&	87.88$\pm$0.1	&	80.8$\pm$0.1	&	89.69$\pm$0.17	&	88.31$\pm$0.09	&	73.42$\pm$0.06	&	76.28$\pm$0.14	&	86.18$\pm$0.16	&	83.27$\pm$0.04	&	80.29$\pm$0.05	&	91.72$\pm$0.25	&	83.95	\\
OVANet	&	78.42$\pm$0.21	&	84.47$\pm$0.43	&	88.3$\pm$0.27	&	80.71$\pm$0.41	&	84.84$\pm$0.45	&	86.53$\pm$0.11	&	64.85$\pm$0.7	&	60.25$\pm$0.53	&	84.74$\pm$0.22	&	82.66$\pm$0.17	&	77.01$\pm$0.2	&	90.66$\pm$0.49	&	80.29	\\
UniOT	&	\textbf{84.19}$\pm$0.27	&	\textbf{90.91}$\pm$0.95	&	\textbf{90.48}$\pm$0.58	&	83.97$\pm$0.52	&	\textbf{91.71}$\pm$0.97	&	88.47$\pm$0.82	&	80.06$\pm$1.1	&	\textbf{82.36}$\pm$0.62	&	87.92$\pm$0.26	&	85.26$\pm$0.45	&	\textbf{82.58}$\pm$0.95	&	\textbf{91.76}$\pm$0.63	&	\textbf{86.64}	\\ \hline
WiSE-FT	&	64.63$\pm$0.04	&	82.12$\pm$0.16	&	84.11$\pm$0.03	&	60.82$\pm$0.31	&	82.93$\pm$0.25	&	79.64$\pm$0.26	&	59.95$\pm$0.38	&	54.5$\pm$0.31	&	80.59$\pm$0.1	&	71.29$\pm$0.23	&	69.77$\pm$0.27	&	90.9$\pm$0.22	&	73.44	\\
CLIP cross-model	&	77.8$\pm$0.06	&	84.46$\pm$0.06	&	87.64$\pm$0.08	&	84.68$\pm$0.1	&	87.19$\pm$0.23	&	88.59$\pm$0.09	&	81.77$\pm$0.11	&	81.85$\pm$0.1	&	89.5$\pm$0.09	&	84.61$\pm$0.09	&	79.31$\pm$0.15	&	83.66$\pm$0.1	&	84.26	\\
\bottomrule
    \end{tabular}
    \end{adjustbox}
    \caption{OfficeHome: CLIP \& (15/0) setting}
    \label{tab:officehome-clip-15-0}
\end{table}

\begin{table}[htp]
    \centering
    \begin{adjustbox}{width=1.0\textwidth,center}
    \begin{tabular}{c|cccccc|c} \toprule
Methods	&	P2R	&	P2S	&	R2P	&	R2S	&	S2P	&	S2R	&	Avg	\\ \midrule
\multicolumn{8}{c}{H-score} \\ \midrule
SO	&	73.95$\pm$0.01	&	70.14$\pm$0.02	&	68.96$\pm$0.11	&	69.83$\pm$0.05	&	66.51$\pm$0.03	&	72.74$\pm$0.02	&	70.36	\\
DANCE	&	74.04$\pm$0.09	&	70.33$\pm$0.05	&	68.79$\pm$0.06	&	70.63$\pm$0.07	&	67.03$\pm$0.11	&	72.22$\pm$0.14	&	70.51	\\
OVANet	&	54.47$\pm$0.44	&	67.84$\pm$0.01	&	67.47$\pm$0.02	&	66.35$\pm$0.17	&	54.74$\pm$0.2	&	34.07$\pm$0.72	&	57.49	\\
UniOT	&	\textbf{77.1}$\pm$0.28	&	\textbf{71.91}$\pm$0.58	&	\textbf{70.0}$\pm$0.24	&	\textbf{72.78}$\pm$0.22	&	\textbf{68.73}$\pm$0.09	&	\textbf{75.3}$\pm$0.43	&	\textbf{72.64}	\\ \midrule

\multicolumn{8}{c}{H$^3$-score} \\ \midrule
SO	&	77.38$\pm$0.01	&	70.06$\pm$0.01	&	\textbf{70.57}$\pm$0.08	&	69.86$\pm$0.03	&	68.85$\pm$0.02	&	76.48$\pm$0.01	&	72.2	\\
DANCE	&	77.44$\pm$0.07	&	\textbf{70.19}$\pm$0.03	&	70.45$\pm$0.04	&	70.39$\pm$0.04	&	\textbf{69.22}$\pm$0.07	&	76.1$\pm$0.1	&	\textbf{72.3}	\\
OVANet	&	61.92$\pm$0.38	&	68.51$\pm$0.01	&	69.53$\pm$0.01	&	67.49$\pm$0.12	&	59.95$\pm$0.16	&	42.6$\pm$0.75	&	61.67	\\
UniOT	&	\textbf{77.62}$\pm$0.21	&	69.74$\pm$0.31	&	69.75$\pm$0.15	&	\textbf{70.58}$\pm$0.04	&	68.8$\pm$0.05	&	\textbf{76.51}$\pm$0.3	&	72.17	\\ \midrule

\multicolumn{8}{c}{UCR} \\ \midrule
SO	&	70.72$\pm$0.02	&	59.82$\pm$0.13	&	61.89$\pm$0.08	&	60.51$\pm$0.15	&	60.66$\pm$0.12	&	70.4$\pm$0.02	&	64.0	\\
DANCE	&	\textbf{71.43}$\pm$0.01	&	\textbf{62.13}$\pm$0.17	&	\textbf{63.72}$\pm$0.13	&	\textbf{63.89}$\pm$0.03	&	\textbf{63.24}$\pm$0.23	&	70.69$\pm$0.07	&	\textbf{65.85}	\\
OVANet	&	68.15$\pm$0.11	&	55.82$\pm$0.17	&	59.94$\pm$0.16	&	56.97$\pm$0.11	&	58.49$\pm$0.14	&	68.39$\pm$0.15	&	61.29	\\
UniOT	&	70.9$\pm$0.21	&	59.07$\pm$0.27	&	60.27$\pm$0.08	&	61.74$\pm$0.25	&	58.76$\pm$0.23	&	\textbf{70.9}$\pm$0.6	&	63.61	\\
\bottomrule
    \end{tabular}
    \end{adjustbox}
    \caption{DomainNet: DINOv2 \& (150/0) setting}
    \label{tab:domainnet-dinov2-150-0}
\end{table}

\begin{table}[htp]
    \centering
    \begin{adjustbox}{width=1.0\textwidth,center}
    \begin{tabular}{c|cccccc|c} \toprule
Methods	&	P2R	&	P2S	&	R2P	&	R2S	&	S2P	&	S2R	&	Avg	\\ \midrule
\multicolumn{8}{c}{H-score} \\ \midrule
SO	&	72.15$\pm$0.06	&	62.77$\pm$0.1	&	62.5$\pm$0.19	&	65.55$\pm$0.09	&	56.61$\pm$0.05	&	74.19$\pm$0.03	&	65.63	\\
DANCE	&	72.2$\pm$0.07	&	62.16$\pm$0.2	&	60.81$\pm$0.06	&	65.86$\pm$0.16	&	56.69$\pm$0.01	&	73.75$\pm$0.02	&	65.25	\\
OVANet	&	77.09$\pm$0.17	&	71.64$\pm$0.04	&	69.57$\pm$0.13	&	72.51$\pm$0.1	&	67.04$\pm$0.11	&	76.58$\pm$0.11	&	72.41	\\
UniOT	&	76.78$\pm$0.08	&	70.35$\pm$0.31	&	\textbf{71.84}$\pm$0.05	&	\textbf{73.56}$\pm$0.15	&	\textbf{69.32}$\pm$0.36	&	77.43$\pm$0.19	&	73.21	\\ \hline
WiSE-FT	&	11.76$\pm$0.11	&	4.52$\pm$0.09	&	8.84$\pm$0.11	&	7.73$\pm$0.07	&	3.77$\pm$0.08	&	10.9$\pm$0.13	&	7.92	\\
CLIP cross-model	&	74.9$\pm$0.1	&	64.16$\pm$0.03	&	62.16$\pm$0.06	&	65.96$\pm$0.05	&	58.29$\pm$0.04	&	77.11$\pm$0.08	&	67.1	\\
\bottomrule
    \end{tabular}
    \end{adjustbox}
    \caption{DomainNet: CLIP \& (150/0) setting}
    \label{tab:domainnet-clip-150-0}
\end{table}

\clearpage
\begin{table}[htp]
    \centering
    \begin{adjustbox}{width=1.0\textwidth,center}
    \begin{tabular}{c|cccccc|c} \toprule
Methods	&	A2D	&	A2W	&	D2A	&	D2W	&	W2A	&	W2D	&	Avg	\\ \midrule
\multicolumn{8}{c}{H-score/H$^3$-score} \\ \midrule
SO	&	85.49$\pm$0.06	&	83.35$\pm$0.18	&	\textbf{67.23}$\pm$0.53	&	\textbf{99.38}$\pm$0.1	&	\textbf{68.51}$\pm$0.33	&	\textbf{100.0}$\pm$0.0	&	\textbf{83.99}	\\
DANCE	&	85.23$\pm$0.48	&	\textbf{83.49}$\pm$0.21	&	60.16$\pm$3.33	&	99.28$\pm$0.0	&	62.17$\pm$2.37	&	99.69$\pm$0.0	&	81.67	\\
OVANet	&	\textbf{88.03}$\pm$0.22	&	83.27$\pm$0.34	&	61.73$\pm$1.68	&	96.0$\pm$0.08	&	64.01$\pm$0.49	&	100.0$\pm$0.0	&	82.17	\\
UniOT	&	63.33$\pm$0.94	&	59.42$\pm$1.83	&	51.9$\pm$0.94	&	79.87$\pm$0.64	&	52.81$\pm$0.52	&	85.29$\pm$3.81	&	65.44	\\ \midrule

\multicolumn{8}{c}{UCR} \\ \midrule
SO	&	\textbf{91.97}$\pm$0.16	&	92.49$\pm$0.16	&	77.77$\pm$0.36	&	99.54$\pm$0.12	&	78.89$\pm$0.17	&	\textbf{100.0}$\pm$0.0	&	90.11	\\
DANCE	&	89.49$\pm$0.53	&	90.69$\pm$0.31	&	70.1$\pm$2.32	&	\textbf{99.62}$\pm$0.0	&	72.06$\pm$1.39	&	99.87$\pm$0.09	&	86.97	\\
OVANet	&	91.97$\pm$0.16	&	92.49$\pm$0.06	&	77.78$\pm$0.28	&	99.54$\pm$0.12	&	78.8$\pm$0.28	&	100.0$\pm$0.0	&	90.1	\\
UniOT	&	91.63$\pm$1.23	&	\textbf{92.79}$\pm$0.57	&	\textbf{81.24}$\pm$0.48	&	98.7$\pm$0.24	&	\textbf{81.74}$\pm$0.26	&	100.0$\pm$0.0	&	\textbf{91.02}	\\
\bottomrule
    \end{tabular}
    \end{adjustbox}
    \caption{Office: DINOv2 \& (31/0) setting}
    \label{tab:office-dinov2-31-0}
\end{table}

\begin{table}[htp]
    \centering
    \begin{adjustbox}{width=1.0\textwidth,center}
    \begin{tabular}{c|cccccc|c} \toprule
Methods	&	A2D	&	A2W	&	D2A	&	D2W	&	W2A	&	W2D	&	Avg	\\ \midrule
\multicolumn{8}{c}{H-score/H$^3$-score} \\ \midrule
SO	&	76.99$\pm$0.54	&	71.98$\pm$0.1	&	68.54$\pm$0.19	&	97.64$\pm$0.15	&	66.94$\pm$0.25	&	99.26$\pm$0.0	&	80.22	\\
DANCE	&	73.29$\pm$1.1	&	63.14$\pm$0.16	&	62.46$\pm$1.66	&	\textbf{98.4}$\pm$0.0	&	57.63$\pm$3.83	&	\textbf{99.64}$\pm$0.32	&	75.76	\\
OVANet	&	72.5$\pm$0.73	&	64.35$\pm$0.3	&	58.02$\pm$1.13	&	96.26$\pm$0.05	&	58.07$\pm$0.66	&	98.67$\pm$0.0	&	74.64	\\
UniOT	&	58.03$\pm$2.06	&	56.8$\pm$1.14	&	53.8$\pm$0.89	&	69.88$\pm$1.5	&	53.23$\pm$0.88	&	67.96$\pm$3.44	&	59.95	\\ \hline
WiSE-FT	&	31.89$\pm$0.31	&	25.87$\pm$0.19	&	39.04$\pm$0.14	&	74.62$\pm$0.26	&	36.65$\pm$0.29	&	79.13$\pm$0.09	&	47.87	\\
CLIP cross-model	&	79.87$\pm$0.45	&	77.27$\pm$0.38	&	73.99$\pm$0.08	&	97.46$\pm$0.0	&	72.03$\pm$0.05	&	98.65$\pm$0.12	&	83.21	\\
\bottomrule
    \end{tabular}
    \end{adjustbox}
    \caption{Office: CLIP \& (31/0) setting}
    \label{tab:office-clip-31-0}
\end{table}

\begin{table}[htp]
    \centering
    \begin{adjustbox}{width=1.0\textwidth,center}
    \begin{tabular}{c|cccccccccccc|c} \toprule
Methods	&	A2C	&	A2P	&	A2R	&	C2A	&	C2P	&	C2R	&	P2A	&	P2C	&	P2R	&	R2A	&	R2C	&	R2P	&	Avg	\\ \midrule
\multicolumn{14}{c}{H-score/H$^3$-score} \\ \midrule
SO	&	\textbf{59.36}$\pm$0.1	&	74.0$\pm$0.07	&	77.72$\pm$0.03	&	56.65$\pm$0.05	&	71.85$\pm$0.04	&	68.49$\pm$0.09	&	52.6$\pm$0.16	&	\textbf{51.14}$\pm$0.18	&	73.53$\pm$0.05	&	69.0$\pm$0.06	&	\textbf{59.9}$\pm$0.14	&	86.39$\pm$0.03	&	66.72	\\
DANCE	&	56.97$\pm$0.31	&	67.65$\pm$0.62	&	78.01$\pm$0.1	&	54.79$\pm$0.41	&	69.48$\pm$1.04	&	67.41$\pm$0.2	&	46.53$\pm$0.36	&	44.19$\pm$0.81	&	71.55$\pm$0.41	&	68.64$\pm$0.32	&	59.15$\pm$0.17	&	\textbf{86.97}$\pm$0.29	&	64.28	\\
OVANet	&	54.35$\pm$0.18	&	\textbf{78.3}$\pm$0.09	&	\textbf{83.43}$\pm$0.13	&	\textbf{74.66}$\pm$0.19	&	\textbf{82.64}$\pm$0.08	&	\textbf{84.98}$\pm$0.21	&	\textbf{61.55}$\pm$0.26	&	39.19$\pm$0.14	&	\textbf{80.72}$\pm$0.14	&	\textbf{70.67}$\pm$0.37	&	51.02$\pm$0.22	&	85.63$\pm$0.3	&	\textbf{70.59}	\\
UniOT	&	50.15$\pm$0.13	&	55.39$\pm$0.71	&	63.4$\pm$0.4	&	52.6$\pm$0.6	&	57.93$\pm$0.62	&	59.99$\pm$0.61	&	45.47$\pm$0.49	&	43.98$\pm$0.22	&	59.39$\pm$0.21	&	58.06$\pm$1.11	&	51.69$\pm$0.32	&	71.62$\pm$0.03	&	55.81	\\ \midrule

\multicolumn{14}{c}{UCR} \\ \midrule
SO	&	72.02$\pm$0.03	&	\textbf{86.92}$\pm$0.04	&	88.95$\pm$0.04	&	\textbf{82.24}$\pm$0.19	&	86.94$\pm$0.07	&	87.58$\pm$0.11	&	\textbf{76.03}$\pm$0.15	&	67.13$\pm$0.22	&	86.8$\pm$0.08	&	84.05$\pm$0.09	&	71.87$\pm$0.19	&	92.3$\pm$0.03	&	81.9	\\
\bottomrule
    \end{tabular}
    \end{adjustbox}
    \caption{OfficeHome: DINOv2 \& (65/0) setting}
    \label{tab:officehome-dinov2-65-0}
\end{table}

\begin{table}[htp]
    \centering
    \begin{adjustbox}{width=1.0\textwidth,center}
    \begin{tabular}{c|cccccccccccc|c} \toprule
Methods	&	A2C	&	A2P	&	A2R	&	C2A	&	C2P	&	C2R	&	P2A	&	P2C	&	P2R	&	R2A	&	R2C	&	R2P	&	Avg	\\ \midrule
\multicolumn{14}{c}{H-score/H$^3$-score} \\ \midrule
SO	&	42.18$\pm$0.02	&	65.78$\pm$0.1	&	67.64$\pm$0.01	&	46.06$\pm$0.18	&	68.16$\pm$0.16	&	64.49$\pm$0.08	&	42.68$\pm$0.12	&	37.5$\pm$0.07	&	71.0$\pm$0.07	&	58.19$\pm$0.19	&	48.05$\pm$0.09	&	85.69$\pm$0.05	&	58.12	\\
DANCE	&	39.45$\pm$0.12	&	59.51$\pm$1.16	&	68.27$\pm$0.05	&	45.45$\pm$0.17	&	58.08$\pm$0.31	&	63.0$\pm$0.1	&	36.2$\pm$0.56	&	35.31$\pm$0.1	&	68.49$\pm$0.29	&	58.1$\pm$0.47	&	47.07$\pm$0.06	&	86.12$\pm$0.02	&	55.42	\\
OVANet	&	53.14$\pm$0.16	&	67.63$\pm$0.26	&	77.34$\pm$0.07	&	61.55$\pm$0.37	&	67.28$\pm$0.31	&	73.51$\pm$0.21	&	50.26$\pm$0.2	&	40.58$\pm$0.43	&	75.23$\pm$0.19	&	68.72$\pm$0.14	&	53.28$\pm$0.23	&	87.32$\pm$0.2	&	64.65	\\
UniOT	&	52.4$\pm$0.23	&	54.54$\pm$0.9	&	64.91$\pm$0.76	&	61.16$\pm$0.31	&	66.6$\pm$1.57	&	72.71$\pm$0.73	&	47.04$\pm$0.66	&	47.5$\pm$0.23	&	59.05$\pm$0.61	&	60.14$\pm$0.58	&	54.32$\pm$0.22	&	70.81$\pm$1.06	&	59.27	\\ \hline
WiSE-FT	&	5.98$\pm$0.11	&	14.86$\pm$0.05	&	17.47$\pm$0.04	&	3.65$\pm$0.12	&	9.79$\pm$0.11	&	9.0$\pm$0.13	&	9.35$\pm$0.07	&	5.21$\pm$0.08	&	26.2$\pm$0.02	&	14.31$\pm$0.11	&	8.7$\pm$0.18	&	39.13$\pm$0.07	&	13.64	\\
CLIP cross-model	&	44.97$\pm$0.19	&	73.26$\pm$0.06	&	72.75$\pm$0.08	&	52.76$\pm$0.25	&	77.72$\pm$0.06	&	71.95$\pm$0.09	&	47.85$\pm$0.1	&	41.82$\pm$0.04	&	73.98$\pm$0.09	&	58.98$\pm$0.03	&	49.1$\pm$0.09	&	86.71$\pm$0.03	&	62.65	\\
\bottomrule
    \end{tabular}
    \end{adjustbox}
    \caption{OfficeHome: CLIP \& (65/0) setting}
    \label{tab:officehome-clip-65-0}
\end{table}

\begin{table}[htp]
    \centering
    \begin{adjustbox}{width=1.0\textwidth,center}
    \begin{tabular}{c|cccccc|c} \toprule
Methods	&	P2R	&	P2S	&	R2P	&	R2S	&	S2P	&	S2R	&	Avg	\\ \midrule
\multicolumn{8}{c}{H-score/H$^3$-score} \\ \midrule
SO	&	56.91$\pm$0.08	&	48.36$\pm$0.05	&	50.97$\pm$0.07	&	50.08$\pm$0.08	&	49.25$\pm$0.07	&	54.4$\pm$0.07	&	51.66	\\
DANCE	&	56.7$\pm$0.08	&	48.93$\pm$0.18	&	50.6$\pm$0.05	&	50.78$\pm$0.2	&	49.03$\pm$0.08	&	53.51$\pm$0.31	&	51.59	\\
OVANet	&	\textbf{71.99}$\pm$0.05	&	48.15$\pm$0.19	&	\textbf{52.21}$\pm$0.1	&	45.3$\pm$0.12	&	\textbf{60.43}$\pm$0.02	&	\textbf{75.43}$\pm$0.03	&	\textbf{58.92}	\\
UniOT	&	59.05$\pm$0.17	&	\textbf{52.93}$\pm$0.09	&	50.64$\pm$0.42	&	\textbf{53.44}$\pm$0.14	&	50.33$\pm$0.28	&	57.59$\pm$0.22	&	54.0	\\
\midrule

\multicolumn{8}{c}{UCR} \\ \midrule
SO	&	76.11$\pm$0.13	&	62.71$\pm$0.04	&	67.28$\pm$0.08	&	62.65$\pm$0.13	&	66.89$\pm$0.03	&	78.02$\pm$0.06	&	68.94	\\
DANCE	&	75.62$\pm$0.12	&	\textbf{64.66}$\pm$0.09	&	\textbf{68.6}$\pm$0.12	&	64.89$\pm$0.12	&	\textbf{67.52}$\pm$0.04	&	77.09$\pm$0.06	&	\textbf{69.73}	\\
OVANet	&	\textbf{76.16}$\pm$0.02	&	62.71$\pm$0.07	&	67.28$\pm$0.07	&	62.75$\pm$0.03	&	66.9$\pm$0.03	&	\textbf{78.04}$\pm$0.05	&	68.97	\\
UniOT	&	75.8$\pm$0.09	&	64.49$\pm$0.12	&	66.91$\pm$0.1	&	\textbf{65.24}$\pm$0.08	&	65.56$\pm$0.06	&	76.46$\pm$0.06	&	69.08	\\
\bottomrule
    \end{tabular}
    \end{adjustbox}
    \caption{DomainNet: DINOv2 \& (345/0) setting}
    \label{tab:domainnet-dinov2-345-0}
\end{table}

\begin{table}[htp]
    \centering
    \begin{adjustbox}{width=1.0\textwidth,center}
    \begin{tabular}{c|cccccc|c} \toprule
Methods	&	P2R	&	P2S	&	R2P	&	R2S	&	S2P	&	S2R	&	Avg	\\ \midrule
\multicolumn{8}{c}{H-score/H$^3$-score} \\ \midrule
SO	&	47.32$\pm$0.07	&	31.92$\pm$0.03	&	33.24$\pm$0.06	&	35.42$\pm$0.06	&	30.24$\pm$0.04	&	51.46$\pm$0.04	&	38.27	\\
DANCE	&	47.09$\pm$0.03	&	32.1$\pm$0.11	&	31.62$\pm$0.24	&	33.88$\pm$0.14	&	29.89$\pm$0.11	&	50.78$\pm$0.04	&	37.56	\\
OVANet	&	65.92$\pm$0.1	&	51.21$\pm$0.11	&	49.47$\pm$0.07	&	54.56$\pm$0.05	&	51.87$\pm$0.19	&	70.27$\pm$0.09	&	57.22	\\
UniOT	&	71.17$\pm$0.32	&	\textbf{60.36}$\pm$0.17	&	\textbf{57.09}$\pm$0.35	&	\textbf{63.79}$\pm$0.17	&	\textbf{56.98}$\pm$0.15	&	72.93$\pm$0.07	&	\textbf{63.72}	\\ \hline
WiSE-FT	&	0.56$\pm$0.01	&	0.09$\pm$0.0	&	0.47$\pm$0.03	&	0.25$\pm$0.01	&	0.11$\pm$0.01	&	0.3$\pm$0.01	&	0.3	\\
CLIP cross-model	&	46.33$\pm$0.11	&	29.96$\pm$0.09	&	29.69$\pm$0.12	&	32.42$\pm$0.1	&	28.35$\pm$0.06	&	50.46$\pm$0.06	&	36.2	\\
\bottomrule
    \end{tabular}
    \end{adjustbox}
    \caption{DomainNet: CLIP \& (345/0) setting}
    \label{tab:domainnet-clip-345-0}
\end{table}

\begin{table}[htp]
    \centering
    \begin{adjustbox}{width=1.0\textwidth,center}
    \begin{tabular}{c|cccccc|c} \toprule
Methods	&	A2D	&	A2W	&	D2A	&	D2W	&	W2A	&	W2D	&	Avg	\\ \midrule
\multicolumn{8}{c}{H-score/H$^3$-score} \\ \midrule
SO	&	91.71$\pm$0.21	&	88.29$\pm$0.46	&	80.88$\pm$0.76	&	99.53$\pm$0.33	&	82.58$\pm$0.19	&	\textbf{100.0}$\pm$0.0	&	90.5	\\
DANCE	&	70.75$\pm$4.51	&	79.7$\pm$3.77	&	65.66$\pm$0.08	&	\textbf{100.0}$\pm$0.0	&	65.83$\pm$0.26	&	100.0$\pm$0.0	&	80.32	\\
OVANet	&	\textbf{95.5}$\pm$0.0	&	\textbf{93.82}$\pm$0.37	&	\textbf{82.73}$\pm$0.15	&	97.49$\pm$0.14	&	\textbf{82.6}$\pm$0.21	&	100.0$\pm$0.0	&	\textbf{92.02}	\\
UniOT	&	37.2$\pm$4.35	&	36.71$\pm$1.82	&	40.67$\pm$0.63	&	43.64$\pm$1.61	&	39.16$\pm$1.08	&	48.85$\pm$3.19	&	41.04	\\ \midrule

\multicolumn{8}{c}{UCR} \\ \midrule
SO	&	\textbf{97.03}$\pm$0.3	&	93.56$\pm$0.28	&	86.95$\pm$0.3	&	99.44$\pm$0.32	&	\textbf{90.54}$\pm$0.34	&	\textbf{100.0}$\pm$0.0	&	94.59	\\
DANCE	&	79.83$\pm$2.35	&	87.34$\pm$0.7	&	71.75$\pm$0.44	&	\textbf{100.0}$\pm$0.0	&	72.3$\pm$0.47	&	100.0$\pm$0.0	&	85.2	\\
OVANet	&	97.03$\pm$0.3	&	\textbf{93.79}$\pm$0.32	&	\textbf{87.09}$\pm$0.38	&	99.44$\pm$0.32	&	90.5$\pm$0.31	&	100.0$\pm$0.0	&	\textbf{94.64}	\\
UniOT	&	49.04$\pm$2.9	&	50.85$\pm$1.73	&	61.93$\pm$0.34	&	66.1$\pm$1.27	&	60.65$\pm$0.23	&	70.49$\pm$2.62	&	59.84	\\
\bottomrule
    \end{tabular}
    \end{adjustbox}
    \caption{Office: DINOv2 \& (10/21) setting}
    \label{tab:office-dinov2-10-21}
\end{table}

\begin{table}[htp]
    \centering
    \begin{adjustbox}{width=1.0\textwidth,center}
    \begin{tabular}{c|cccccc|c} \toprule
Methods	&	A2D	&	A2W	&	D2A	&	D2W	&	W2A	&	W2D	&	Avg	\\ \midrule
\multicolumn{8}{c}{H-score/H$^3$-score} \\ \midrule
SO	&	86.4$\pm$0.5	&	83.72$\pm$0.24	&	89.13$\pm$0.52	&	96.43$\pm$0.22	&	84.28$\pm$0.55	&	\textbf{98.75}$\pm$0.0	&	89.79	\\
DANCE	&	65.23$\pm$6.21	&	43.33$\pm$4.71	&	57.13$\pm$8.0	&	90.0$\pm$0.0	&	47.77$\pm$7.84	&	97.5$\pm$0.0	&	66.83	\\
OVANet	&	91.42$\pm$0.2	&	77.48$\pm$2.74	&	81.96$\pm$4.4	&	95.79$\pm$0.11	&	81.63$\pm$4.31	&	96.92$\pm$0.0	&	87.53	\\
UniOT	&	43.36$\pm$1.88	&	36.25$\pm$2.27	&	39.29$\pm$1.02	&	42.39$\pm$0.83	&	37.23$\pm$2.93	&	49.34$\pm$1.09	&	41.31	\\ \hline
WiSE-FT	&	39.89$\pm$0.39	&	28.34$\pm$0.15	&	50.65$\pm$0.36	&	76.36$\pm$0.18	&	43.01$\pm$0.44	&	83.15$\pm$0.39	&	53.57	\\
CLIP cross-model	&	89.62$\pm$0.2	&	87.18$\pm$0.18	&	93.65$\pm$0.13	&	\textbf{96.64}$\pm$0.0	&	91.03$\pm$0.13	&	97.19$\pm$0.2	&	92.55	\\
\bottomrule
    \end{tabular}
    \end{adjustbox}
    \caption{Office: CLIP \& (10/21) setting}
    \label{tab:office-clip-10-21}
\end{table}

\begin{table}[htp]
    \centering
    \begin{adjustbox}{width=1.0\textwidth,center}
    \begin{tabular}{c|cccccccccccc|c} \toprule
Methods	&	A2C	&	A2P	&	A2R	&	C2A	&	C2P	&	C2R	&	P2A	&	P2C	&	P2R	&	R2A	&	R2C	&	R2P	&	Avg	\\ \midrule
\multicolumn{14}{c}{H-score/H$^3$-score} \\ \midrule
SO	&	\textbf{61.02}$\pm$0.08	&	74.29$\pm$0.1	&	79.21$\pm$0.1	&	50.62$\pm$0.23	&	61.51$\pm$0.16	&	62.24$\pm$0.14	&	49.47$\pm$0.18	&	\textbf{50.7}$\pm$0.39	&	70.06$\pm$0.08	&	69.75$\pm$0.23	&	\textbf{61.87}$\pm$0.37	&	82.23$\pm$0.14	&	64.41	\\
DANCE	&	56.34$\pm$1.01	&	57.05$\pm$1.2	&	79.52$\pm$1.6	&	47.43$\pm$3.36	&	50.13$\pm$1.26	&	57.75$\pm$0.51	&	35.09$\pm$0.43	&	34.65$\pm$0.42	&	63.59$\pm$1.81	&	66.5$\pm$0.6	&	56.41$\pm$0.11	&	79.9$\pm$1.89	&	57.03	\\
OVANet	&	58.72$\pm$0.04	&	\textbf{81.24}$\pm$0.61	&	\textbf{87.33}$\pm$0.36	&	\textbf{72.12}$\pm$0.39	&	\textbf{77.08}$\pm$0.33	&	\textbf{81.4}$\pm$0.33	&	\textbf{63.93}$\pm$0.24	&	41.44$\pm$0.48	&	\textbf{80.21}$\pm$0.48	&	\textbf{74.46}$\pm$0.25	&	54.17$\pm$0.31	&	\textbf{83.33}$\pm$0.22	&	\textbf{71.29}	\\
UniOT	&	43.53$\pm$0.88	&	40.65$\pm$0.59	&	39.34$\pm$1.4	&	34.31$\pm$2.27	&	37.81$\pm$1.26	&	33.29$\pm$0.73	&	32.87$\pm$0.75	&	34.05$\pm$0.74	&	37.52$\pm$0.5	&	44.56$\pm$1.09	&	42.73$\pm$0.26	&	41.94$\pm$1.3	&	38.55	\\ \midrule

\multicolumn{14}{c}{UCR} \\ \midrule
SO	&	75.88$\pm$0.05	&	\textbf{88.85}$\pm$0.16	&	91.2$\pm$0.03	&	\textbf{80.44}$\pm$0.49	&	80.04$\pm$0.07	&	\textbf{84.58}$\pm$0.18	&	\textbf{76.55}$\pm$0.3	&	67.72$\pm$0.41	&	\textbf{85.86}$\pm$0.09	&	85.49$\pm$0.15	&	\textbf{74.43}$\pm$0.1	&	91.07$\pm$0.21	&	\textbf{81.84}	\\
\bottomrule
    \end{tabular}
    \end{adjustbox}
    \caption{OfficeHome: DINOv2 \& (25/40) setting}
    \label{tab:officehome-dinov2-25-40}
\end{table}

\begin{table}[htp]
    \centering
    \begin{adjustbox}{width=1.0\textwidth,center}
    \begin{tabular}{c|cccccccccccc|c} \toprule
Methods	&	A2C	&	A2P	&	A2R	&	C2A	&	C2P	&	C2R	&	P2A	&	P2C	&	P2R	&	R2A	&	R2C	&	R2P	&	Avg	\\ \midrule
\multicolumn{14}{c}{H-score/H$^3$-score} \\ \midrule
SO	&	44.33$\pm$0.19	&	65.34$\pm$0.35	&	71.4$\pm$0.35	&	44.78$\pm$0.1	&	59.45$\pm$0.47	&	63.91$\pm$0.17	&	46.61$\pm$0.26	&	41.37$\pm$0.14	&	72.02$\pm$0.12	&	59.08$\pm$0.17	&	53.28$\pm$0.1	&	78.2$\pm$0.12	&	58.31	\\
DANCE	&	34.87$\pm$0.17	&	41.08$\pm$1.78	&	74.03$\pm$1.62	&	31.93$\pm$1.16	&	34.31$\pm$0.22	&	54.84$\pm$2.92	&	25.85$\pm$1.81	&	21.6$\pm$0.81	&	63.45$\pm$2.65	&	57.55$\pm$0.35	&	45.7$\pm$0.79	&	74.36$\pm$0.06	&	46.63	\\
OVANet	&	60.09$\pm$0.43	&	70.91$\pm$0.3	&	81.93$\pm$0.39	&	60.59$\pm$0.48	&	61.88$\pm$1.41	&	72.19$\pm$0.26	&	52.46$\pm$0.48	&	45.8$\pm$1.0	&	76.8$\pm$0.16	&	68.38$\pm$0.23	&	58.47$\pm$0.2	&	81.48$\pm$0.52	&	65.92	\\
UniOT	&	44.17$\pm$0.19	&	43.91$\pm$0.93	&	39.98$\pm$1.07	&	47.21$\pm$0.58	&	41.66$\pm$2.14	&	47.94$\pm$3.9	&	37.16$\pm$1.5	&	44.8$\pm$0.48	&	39.38$\pm$1.05	&	48.13$\pm$0.6	&	43.92$\pm$0.44	&	44.92$\pm$2.23	&	43.6	\\ \hline
WiSE-FT	&	9.8$\pm$0.04	&	17.69$\pm$0.16	&	22.75$\pm$0.08	&	6.1$\pm$0.04	&	11.27$\pm$0.08	&	13.4$\pm$0.28	&	11.26$\pm$0.09	&	7.26$\pm$0.14	&	29.79$\pm$0.09	&	17.6$\pm$0.12	&	13.3$\pm$0.28	&	38.55$\pm$0.19	&	16.56	\\
CLIP cross-model	&	48.18$\pm$0.04	&	70.76$\pm$0.25	&	75.41$\pm$0.24	&	51.32$\pm$0.17	&	70.49$\pm$0.36	&	70.68$\pm$0.13	&	52.32$\pm$0.19	&	48.38$\pm$0.12	&	75.23$\pm$0.09	&	59.69$\pm$0.2	&	55.33$\pm$0.29	&	79.92$\pm$0.18	&	63.14	\\
\bottomrule
    \end{tabular}
    \end{adjustbox}
    \caption{OfficeHome: CLIP \& (25/40) setting}
    \label{tab:officehome-clip-25-40}
\end{table}

\begin{table}[htp]
    \centering
    \begin{adjustbox}{width=1.0\textwidth,center}
    \begin{tabular}{c|cccccc|c} \toprule
Methods	&	P2R	&	P2S	&	R2P	&	R2S	&	S2P	&	S2R	&	Avg	\\ \midrule
\multicolumn{8}{c}{H-score/H$^3$-score} \\ \midrule
SO	&	54.81$\pm$0.08	&	49.95$\pm$0.11	&	50.3$\pm$0.28	&	50.85$\pm$0.13	&	45.97$\pm$0.05	&	51.3$\pm$0.02	&	50.53	\\
DANCE	&	53.56$\pm$0.24	&	\textbf{51.09}$\pm$0.37	&	49.55$\pm$0.48	&	\textbf{50.97}$\pm$0.82	&	45.88$\pm$0.27	&	44.77$\pm$0.81	&	49.3	\\
OVANet	&	\textbf{70.74}$\pm$0.07	&	49.81$\pm$0.17	&	\textbf{52.21}$\pm$0.11	&	45.79$\pm$0.16	&	\textbf{59.03}$\pm$0.03	&	\textbf{73.48}$\pm$0.07	&	\textbf{58.51}	\\
UniOT	&	46.39$\pm$0.55	&	47.44$\pm$0.16	&	44.6$\pm$0.28	&	48.12$\pm$0.55	&	40.24$\pm$0.3	&	43.68$\pm$0.36	&	45.08	\\
\midrule

\multicolumn{8}{c}{UCR} \\ \midrule
SO	&	73.96$\pm$0.08	&	63.8$\pm$0.13	&	66.12$\pm$0.22	&	65.89$\pm$0.35	&	67.25$\pm$0.02	&	\textbf{75.39}$\pm$0.05	&	68.73	\\
DANCE	&	73.47$\pm$0.13	&	\textbf{66.02}$\pm$0.1	&	\textbf{66.62}$\pm$0.5	&	\textbf{66.0}$\pm$0.82	&	\textbf{67.49}$\pm$0.27	&	72.03$\pm$0.19	&	68.61	\\
OVANet	&	\textbf{74.01}$\pm$0.05	&	63.81$\pm$0.08	&	66.26$\pm$0.05	&	65.84$\pm$0.07	&	67.14$\pm$0.03	&	75.38$\pm$0.03	&	\textbf{68.74}	\\
UniOT	&	66.45$\pm$0.17	&	59.91$\pm$0.09	&	57.66$\pm$0.25	&	60.4$\pm$0.34	&	54.19$\pm$0.51	&	63.7$\pm$0.06	&	60.38	\\
\bottomrule
    \end{tabular}
    \end{adjustbox}
    \caption{DomainNet: DINOv2 \& (150/195) setting}
    \label{tab:domainnet-dinov2-150-195}
\end{table}

\begin{table}[htp]
    \centering
    \begin{adjustbox}{width=1.0\textwidth,center}
    \begin{tabular}{c|cccccc|c} \toprule
Methods	&	P2R	&	P2S	&	R2P	&	R2S	&	S2P	&	S2R	&	Avg	\\ \midrule
\multicolumn{8}{c}{H-score/H$^3$-score} \\ \midrule
SO	&	44.52$\pm$0.09	&	31.86$\pm$0.16	&	32.39$\pm$0.02	&	34.58$\pm$0.12	&	25.71$\pm$0.11	&	46.24$\pm$0.26	&	35.88	\\
DANCE	&	42.13$\pm$0.74	&	28.66$\pm$0.23	&	25.54$\pm$0.33	&	27.51$\pm$0.41	&	21.75$\pm$0.42	&	39.93$\pm$0.71	&	30.92	\\
OVANet	&	65.01$\pm$0.07	&	51.31$\pm$0.26	&	48.81$\pm$0.22	&	54.76$\pm$0.22	&	47.84$\pm$0.25	&	67.4$\pm$0.19	&	55.85	\\
UniOT	&	60.68$\pm$0.59	&	54.95$\pm$0.29	&	50.63$\pm$0.31	&	56.8$\pm$0.05	&	47.35$\pm$0.97	&	60.69$\pm$0.56	&	55.18	\\ \hline
WiSE-FT	&	0.57$\pm$0.01	&	0.06$\pm$0.01	&	0.52$\pm$0.04	&	0.32$\pm$0.02	&	0.07$\pm$0.01	&	0.2$\pm$0.01	&	0.29	\\
CLIP cross-model	&	43.44$\pm$0.12	&	29.73$\pm$0.18	&	28.59$\pm$0.05	&	31.68$\pm$0.23	&	24.92$\pm$0.05	&	46.01$\pm$0.03	&	34.06	\\
\bottomrule
    \end{tabular}
    \end{adjustbox}
    \caption{DomainNet: CLIP \& (150/195) setting}
    \label{tab:domainnet-clip-150-195}
\end{table}


\clearpage

\bibliographystyle{plain}
\bibliography{bing}

\end{document}